\documentclass[12pt]{article}
\usepackage{amsmath}
\usepackage{amsmath,amssymb}
\usepackage{amsthm}
\usepackage{graphicx}
\usepackage{enumerate}
\usepackage{natbib}
\usepackage{url} 
\usepackage{mathrsfs}
\usepackage{dsfont}
\usepackage{bbm}
\usepackage{amsthm}
\usepackage{mathtools}
\usepackage{booktabs}
\usepackage{amsfonts}
\usepackage{nicefrac}
\usepackage{microtype}
\usepackage{xcolor}
\usepackage{lipsum}
\usepackage{float}
\usepackage{tabularx}
\usepackage{subcaption}
\usepackage{multicol}
\usepackage{multirow}
\usepackage{bm}
\usepackage{mathtools}
\usepackage{mathrsfs}
\usepackage{dsfont}
\usepackage{amsthm}
\usepackage{appendix}

\newtheorem{theorem}{Theorem}[section]

\newtheorem{lemma}[theorem]{Lemma}
\newtheorem{assumption}{Assumption}

\newcommand{\indep}{\rotatebox[origin=c]{90}{$\models$}}
\newcommand{\probP}{\text{I\kern-0.15em P}}

\newcommand{\blind}{0}

\begin{document}

\def\spacingset#1{\renewcommand{\baselinestretch}%
{#1}\small\normalsize} \spacingset{1}


\if0\blind
{
  \title{\bf CausalEGM: a general causal inference framework by encoding generative modeling}
  \author{Qiao Liu\dag, Zhongren Chen\dag, Wing Hung Wong\thanks{Corresponding Author. \dag Co-first Authors. 
}\hspace{.2cm}\\
    Department of Statistics, Stanford University\\}
  \maketitle
} \fi

\if1\blind
{
  \bigskip
  \bigskip
  \bigskip
  \begin{center}
    {\LARGE\bf CausalEGM: a general causal inference framework by encoding generative modeling}
\end{center}
  \medskip
} \fi

\bigskip
\begin{abstract}
Although understanding and characterizing causal effects have become essential in observational studies, it is challenging when the covariates are high-dimensional. In this article, we develop a general framework \textit{CausalEGM} for estimating causal effects by encoding generative modeling, which can be applied in both binary and continuous treatment settings. Under the potential outcome framework with unconfoundedness, we establish a bidirectional transformation between the high-dimensional covariate space and a low-dimensional latent space where the density is known (e.g., multivariate normal distribution). Through this, CausalEGM simultaneously decouples the dependencies of covariates on both treatment and outcome and maps the covariates to the low-dimensional latent space. By conditioning on the low-dimensional latent features, CausalEGM can estimate the causal effect for each individual or the average causal effect within a population. Our theoretical analysis shows that the excess risk for CausalEGM can be bounded through empirical process theory. Under an assumption on encoder-decoder networks, the consistency of the estimate can be guaranteed. In a series of experiments, CausalEGM demonstrates superior performance over existing methods for both binary and continuous treatments. Specifically, we find CausalEGM to be substantially more powerful than competing methods in the presence of large sample sizes and high-dimensional covariates. The software of CausalEGM is freely available at \url{https://github.com/SUwonglab/CausalEGM}.

\end{abstract}

\noindent%
\vfill

\newpage
\spacingset{1.9} 
\section{Introduction}
\label{sec:intro}

Given the observational data, drawing inferences about the causal effect of a treatment is crucial to many scientific and engineering problems and attracts immense interest in a wide variety of areas. For example, (1) ~\cite{zhang2017mining} investigated the effect of a drug on health outcomes in personalized medicine; (2) ~\cite{panizza2014public} evaluated the effectiveness of public policies from the governments; (3) ~\cite{kohavi2017online} conducted A/B tests to select a better recommendation strategy by a commercial company. Historically, the small sample size of many datasets imposes an impediment to meaningfully exploring the treatment effect by traditional subgroup analysis. In the big data era, there has been an explosion of data accumulation. We, therefore, require more powerful tools for accurate estimates of causal effects from large-scale observational data.

Researchers are more interested in learning causation than correlation in causal inference. The most effective way to learn the causality is to conduct a randomized controlled trial (RCT), in which subjects are randomly assigned to an experimental group receiving the treatment/intervention and a control group for comparison. Then the difference between the experimental and control group of the outcome measures the efficacy of treatment/intervention. RCT has become the golden standard in studying causal relationships, as randomization can potentially limit all sorts of bias. However, RCT is time-consuming, expensive, and problematic with generalisability (participants in RCT are not always representative of their demographic). In contrast, observational studies can provide valuable evidence and examine effects in “real world” settings, while RCT tends to evaluate treatment effects under ideal conditions among highly selected populations. Given the observational data, we know each individual's treatment, outcome, and covariates. The mechanism of how treatment causally affects the outcome needs to be discovered. One goal is to estimate the counterfactual outcomes. For example, ``would this patient have different health status if he/she received a different therapy?" In real-world applications, treatments are typically not assigned at random due to the selection bias introduced by confounders. The treated population may thus differ significantly from the general population. Accurate estimation of the causal effect involves dealing with confounders, which are variables that affect both treatment and outcome. Failing to adjust for confounding effect may lead to biased estimates and wrong conclusions. 

Many frameworks have been proposed to solve the above problems. The potential outcome model in~\cite{rubin1974estimating} and ~\cite{splawa1990application}, also known as the Neyman–Rubin causal model, is arguably the most widely used framework. It makes precise reasoning about causation and the underlying assumptions. To measure the causal effect of a treatment, we need to compare the factual and counterfactual outcomes of each individual. As it is impossible to observe the potential outcomes of the same individual under different treatment conditions, the inference task can be viewed as a ``missing data" problem where the counterfactual outcome needs to be estimated. Once we solve the ``missing data" problem at an individual or a population-average level, the corresponding individual causal effect or average causal effect can be estimated.

Classic methods of non-parametric estimation of causal effect under the potential outcome framework include re-weighting, matching, and stratification, see the review article~\cite{imbens2004nonparametric} in detail. These methods often perform well when the dimension of covariates is low, but break down when the number of covariates is large. In recent years, the prosperity of machine learning has largely accelerated the development of causal inference algorithms. In this article, we explore the advances in machine learning, especially deep learning, for improving the performance in causal effect estimation. Specifically, we explored how to apply deep generative methods to map the high-dimensional covariates to a latent space with a desired distribution. The proposed dimension reduction scheme enables conditioning on the low-dimensional latent features, which provides new insights into handling the high-dimensional covariates.

\subsection{Related works}
\label{sec:rela}
Our work contributes to the literature on estimating causal effects using deep generative models. Most of the works in this field are under binary treatment settings. For example, re-weighting methods, such as IPW from \cite{rosenbaum1987model}, and \cite{robins1994estimation} assign appropriate weight to each unit to eliminate selection bias. Matching-based methods provide a solution to directly compare the outcomes between the treated and control group within the matched samples. A detailed review of matching methods can be found in \cite{stuart2010matching}. Another type of popular method in causal inference is based on the decision tree. These tree-based methods use non-parametric classification or regression by learning decision rules inferred from data. See \cite{athey2016recursive}, \cite{hill2011bayesian} and \cite{wager2018estimation}. Recently, neural networks have been applied to causal inference, which demonstrates compelling and promising results. See \cite{shalit2017estimating}, \cite{shi2019adapting}, \cite{louizos2017causal}, and \cite{yoon2018ganite}. Most of these efforts are under the binary treatment setting. There are some limitations of these approaches. First, these models typically utilize separate networks for estimating the outcome function under different treatment conditions. Such treatment-specific networks are difficult to be generalized to continuous treatment. Second, those neural network-based methods focus on minimizing the prediction error for the counterfactual outcome while lacking enough theoretical analysis to explain the rationality for the model design and architecture.

As for methods that deal with continuous treatment, a lot of efforts are focused on developing the theory of generalized propensity score from \cite{hirano2004propensity}. See the doubly robust estimators \cite{robins2001comment}, the tree-based method \cite{hill2011bayesian}, \cite{lee2018partial}, and \cite{galagate2016causal} for other regression-based models. There are also non-parametric methods that do not require the correct specification of the models that relate the treatment or outcome to the covariates. See \cite{flores2007estimation}, \cite{kennedy2017non}, \cite{fong2018covariate} and \cite{colangelo2020double}. However, most of the regression-based methods require restrictive conditions on the relationship between covariates and treatment or outcome. For example, \cite{galagate2016causal} only considers the case when the average dose-response function (ADRF) is quadratic. \cite{fong2018covariate} relies on the assumption that the treatment has a linear relationship with covariates. Such a strong assumption hinders the wide application of these methods. Empirically, many of these methods fail under the presence of high dimensional covariates and cannot scale to large-scale datasets. 

To overcome the above limitations, we develop CausalEGM, a general framework for estimating the treatment effect using encoding generative modeling. The CausalEGM model differs from existing methods in the following aspects. 1) Instead of using treatment-specific networks, CausalEGM exploits a uniform model architecture, which is applicable to both discrete and continuous treatment settings. 2) CausalEGM imposes an encoding-generative dimension reduction scheme to decouple the dependency of covariates on treatment and outcome while most existing methods fail to distinguish the dependencies. 3) CausalEGM does not assume any pre-specification treatment model and the outcome model. 
To sum up, the main contribution of this article is to propose a new framework to map high-dimensional covariates to low-dimensional latent features by an encoding-generating scheme. Through this, the latent features with a desired distribution using adversarial training make it easy to condition on. The unified model design also enables the treatment effect estimation under both binary and continuous treatment settings. A series of systematical experiments on benchmark datasets demonstrate that our framework outperforms state-of-the-art methods under various settings.

\section{Method}
\label{sec:meth}
\subsection{Problem Formulation}
We are interested in the causal effect of a variable $X$ on another variable $Y$. $X$ is usually called the treatment (or exposure) variable. $Y$ is called the response (or outcome) variable. We assume $Y$ is real-valued and $X\in \mathscr{X}$ where $\mathscr{X}$ is either a finite set or a bounded interval in $\mathbb{R}$. $X$ and $Y$ are related by a deterministic outcome equation $Y=f(X, V,\epsilon)$ where $V$ represents an observed multi-dimensional covariate, $\epsilon$ represents the set of all other (unobserved) variables that may affect $X$ and $Y$. Conceptually, $(Y, X, V, \epsilon) = (Y, X, V, \epsilon)(\omega)$ is a random variable whose value depends on the sampling unit $\omega$ in an underlying sample space $\Omega$. We observe $(Y, X, V)(\omega_i)$ where $\{\omega_i|i=1,...,n\}$ are i.i.d samples drawn from $\Omega$. The outcome equation $f$ is unknown or assumed to belong to a very general class of functions.

To investigate causal effects, we assume that for each sampling unit $\omega$, there is a set of “potential outcomes” for $\{Y(x)(\omega) = f(x, V(\omega), \epsilon(\omega)), x\in \mathscr{X}\}$ that are potentially measurable. For each sampling unit $\omega$, how the outcome will respond to changing treatment is given by the function $Y(\cdot)(\omega):\mathscr{X}\rightarrow \mathbb{R}$. Thus, these unit-specific response functions capture the causal relations of interest. The main goal of this paper is to estimate their population average.
$$
\mu(x)=\mathbb{E}(Y(x))=\mathbb{E}(f(x,V,\epsilon))
$$
where the function $\mu(x)$ is known as the average dose-response function (ADRF) if $X$ is continuous.

Since we only observe the potential outcome selected by the treatment variable $X(\omega)$, i.e., $Y(\omega)=f(X(\omega), V(\omega), \epsilon(\omega))$, the random variable $Y(x)$ is not directly observable, and its expectation $\mu(x)$ is generally not identifiable from the joint distribution of the observed $(Y,X,V)$. Additional assumptions are needed for the identification of $\mu(x)$. Since $Y(x)$ is a deterministic function of $V$ and $\epsilon$, an “unconfoundedness” condition is imposed, which requires that $\epsilon$ and $X$ are independent conditional on $V$. In other words, once $V$ is given, there should be no unobserved confounding variables that drive correlated changes in the exposure and the outcome.

\begin{assumption}\label{as:unconfoundedness}
(\textbf{unconfoundedness})
Condition on $V$, the treatment $X$, is independent of $\epsilon$,
$$X\indep \epsilon|V.$$
\end{assumption}

Note that the above assumption is a bit different from the well-known "unconfoundedness" assumption where $X\indep Y(x)|V$. The assumption~\ref{as:unconfoundedness} implies that the well-known "unconfoundedness" assumption $X\indep Y(x)|V$ still holds as $Y(x)$ is a deterministic function of $V$ and $\epsilon$ and $Y(x)$ is determined only by $\epsilon$ conditional on $V$. It is known that under the conditional independence assumption, the ADRF is identifiable via the equation.

\begin{eqnarray*}
\mu(x)=\int \mathbb{E}(Y|X=x,V=v)p_{V}(v)dv 
\end{eqnarray*}
where $p_{V}(\cdot)$ is the marginal density of the $V$. If $V$ is a low-dimensional variable, this suggests that we can estimate $\mu(x)$ by the empirical expectation $\mathbb{E}_{n}(t(x,V))$ where $t(x,v)$ is an estimate of $\mathbb{E}(Y|X=x,V=v)$ obtained by nonparametric regression of $Y$ on $X$ and $V$. However, to ensure unconfoundedness, $V$ should include all confounding covariates that can potentially affect both $X$ and $Y$. Thus, in many applications, we must deal with a high-dimensional covariate $V$. This makes the method unattractive as high-dimensional nonparametric regression is difficult in general. Furthermore, unlike the variables in $V$, the exposure variable $X$ is a key variable that should be given special consideration, which is not the case for most non-parametric regression methods. This inherent tension between the size of $V$ and the feasibility of nonparametric regression makes it difficult to use the above equation for estimating $\mu$.

To deal with this tension, in this paper we assume a modified version of unconfoundedness:

\begin{assumption}\label{as:modified unconfounderness}   

There exists a low dimensional feature $Z_0=Z_0(V)$, which can be extracted from the high dimensional covariate $V$ so that $\epsilon$ and $V$ are independent of $X$ conditional on $Z_0$. 

\end{assumption}

Under Assumption \ref{as:modified unconfounderness}, we have
\begin{lemma}
\begin{equation} \label{eqn:1}
\mu(x)=\int \mathbb{E}(Y|X=x,Z_0=z_0)p_{Z_0}(z_0)dz_0  
\end{equation}
\label{thm:identifiable}
\end{lemma}
Since $Z_0$ is of low dimension, it is easy to use (\ref{eqn:1}) provided we know $Z_0$ as a function of $V$. Thus, the causal inference problem is transformed into the problem of learning $Z_0(V)$ from the i.i.d. sample ${(Y_i,X_i,V_i) = (Y,X,V)(S_i):i= 1,2,...,n}$. To learn this function, we propose a deep generative approach, which we call “Encoding Generative Modeling”, that allows simultaneous learning of an encoder for the high-dimensional $V$ and a generative model for $(Y,X,V)$. By imposing a suitable constraint on the generative model, one can ensure that certain subsets of the features computed by the encoder can be used as the low-dimensional feature $Z_0$ in the above condition. The conditional expectation in equation (\ref{eqn:1}) is approximated by a neural network (F network). In practice, we use $\hat{\mu}(x)=\frac{1}{n}\sum_{i=1}^{n}F(X=x,Z_0=z_0^{(i)},Z_1=z_1^{(i)})$ for estimating the dose-response function where $n$ is the sample size. In binary treatment settings, the counterfactual outcome for the $i^{th}$ sample is estimated as $y_{CF}^{(i)}=F(X=1-x^{(i)},Z_0=z_0^{(i)},Z_1=z_1^{(i)})$.

\begin{figure}[htbp]
  \centering
  \includegraphics[width=0.6\columnwidth]{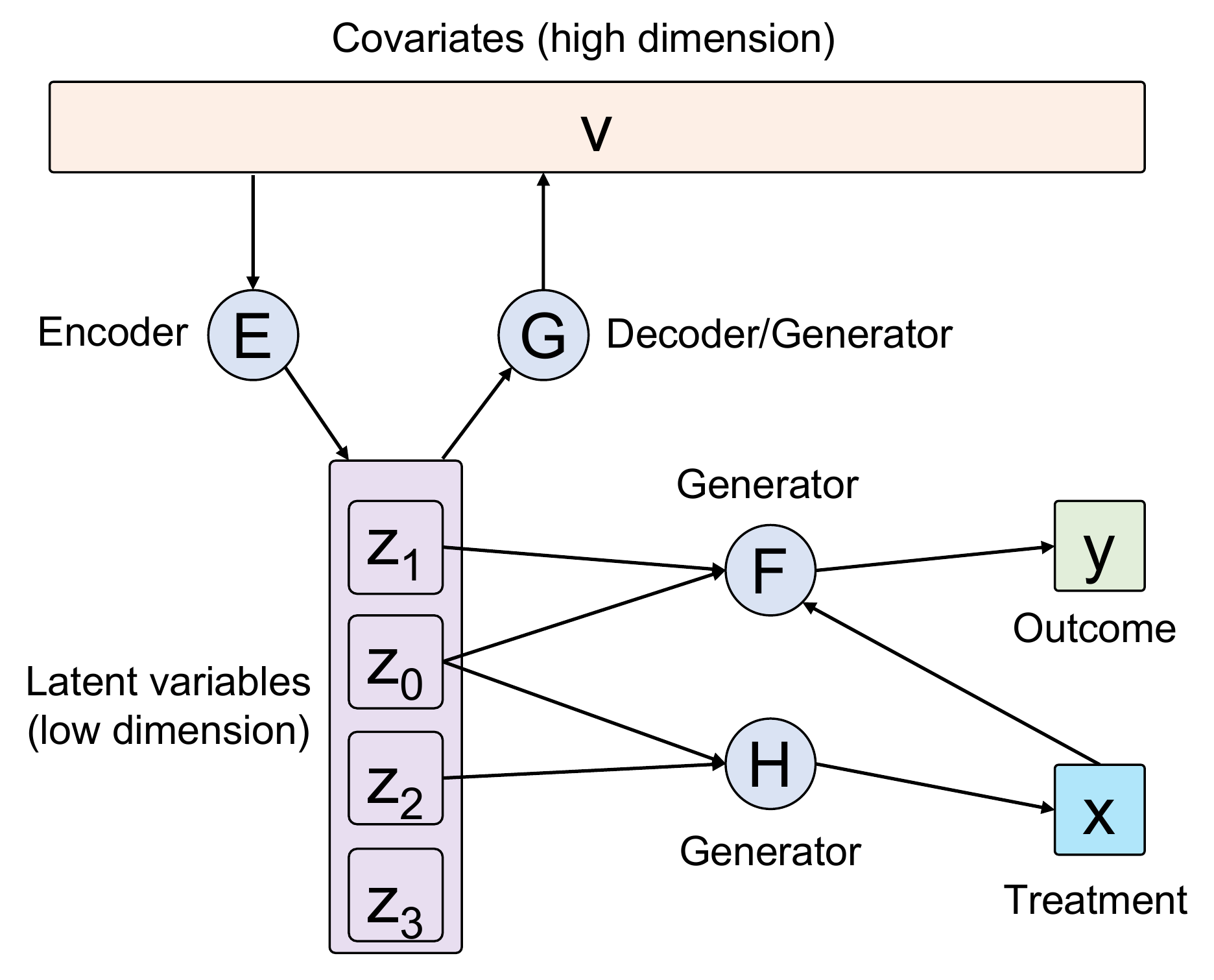}
  \caption{The overview of CausalEGM model. Variables are in rectangles. Functions are in circles, with incoming arrows indicating inputs to the function and outgoing arrows indicating outputs of the function. Each function is modeled by a neural network. CausalEGM takes triplets of (X,Y,V) as input. (E, G) networks are used for mapping the covariates into a latent space. F and H networks are used for recovering the outcome and treatment, respectively. As G, F, and H networks take latent variables as input(s), which have a desired distribution, they are also known as generator networks.}
  \label{fig:model_overview}
\end{figure}

\subsection{An Encoding Generative Model for Causal Inference}
Our model is described in Figure~\ref{fig:model_overview}. To handle the high dimension of $V$, we embed $V$ into a low-dimensional latent space using an encoder function $Z=E(V)$ and a generator$/$decoder function $V=G(Z)$. Note that such bidirectional transformation is inspired by a previous work called Roundtrip for density estimation \citep{liu2021density}. In a standard autoencoder, these functions are learned by minimizing the reconstruction error between $G(E(V))$ and $V$ over the observed sample of $V$. Here, however, we will also impose a “distribution-matching” objective in addition to the reconstruction error. Specifically, assuming a pre-specified distribution for $Z,$ we also want the distribution of $V^{\star}=G(Z)$ to match the distribution of $V$. In this paper, the distribution for $Z$ is assumed to be a standard Gaussian vector. Furthermore, there is a reconstruction objective for $E(G(Z))$ and a distribution matching objective for $E(V)$ in the latent space. We use deep neural networks to represent the functions $E()$ and $G()$. This part of our model, which deals with the relation between $V$ and $Z$, can be viewed as an autoencoder for $V$ where the decoder also serves as a generating function of a generative adversarial network (GAN). Following the common practice in the GAN literature, we use an adversarial loss (i.e., maximizing the classification power between the generated and observed data) in addition to the reconstruction loss in the training. However, the learning of $E()$ and $G()$ should not be based on $V$ alone. Rather, they must be coupled with the learning of generative models for $X$ and $Y$, which are the variables of interest in the causal inference. To do this, we assume that the feature vector $Z=E(V)$ can be partitioned into different sub-vectors that have different roles in the generators for $X$ and $Y$. Specifically, $Z=(Z_0,Z_1,Z_2,Z_3)$, $Y=F(X,Z_0,Z_1)+\epsilon_1$ and $X=H(Z_0,Z_2)+\epsilon_2$. These relations are depicted in the lower half of Figure~\ref{fig:model_overview}, where for clarity, the independent noises $\epsilon_1$ and $\epsilon_2$ are not shown in the input to the generators. Conceptually, $Z_0$ represents the covariate features that affect both treatment and outcome, $Z_1$ represents the covariate features that affect only treatment, $Z_2$ represents the covariate features that affect only the outcome, and $Z_3$ represents the remaining covariate features that are also important for the representation of $V$. CausalEGM is highly flexible for handling different treatment settings. We just need to adjust the activation function in the last layer of the $H$ network for different types of treatment.

\subsection{Model training}
The CausalEGM model consists of a bidirectional transformation module and two feed-forward neural networks. The bidirectional transformation module is used to project the covariates to a low-dimensional space and decouple the dependencies. This bidirectional module is composed of two generative adversarial networks (GANs). In one direction, the encoder network $E$ aims to transform the covariates into latent features, whose distribution matches the standard multivariate Gaussian distribution. A discriminator $D_z$ network tries to distinguish data sampled from the multivariate Gaussian distribution (labeled as positive one) from data generated by the $E$ network (labeled as zero). Similarly, there is another discriminator network in the GAN model that works in the reverse direction, where the generator/decoder network $G$ transforms the latent feature back to the original covariate space to match the empirical distribution for the covariate. A discriminator network $D$ can be considered as a binary classifier where $D(x)=1$ for latent multivariate normal and $D(x)=0$ for the distribution induced by the encoder from the empirical data distribution. We use WGAN-GP \citep{WGAN-GP} as the architecture for the GAN implementation, where the gradient penalty of discriminators is considered as an additional loss term. Thus, the loss function of the adversarial training for distribution matching in latent space has two terms

\begin{equation}
\left\{
\begin{aligned}
\mathcal{L}_{GAN}(E) =& -\mathop{\mathbb{E}}_{v\sim p_{emp}(v)}[D_{z,-1}(E(v))]\\ \mathcal{L}_{GAN}(D_z) =& -\mathop{\mathbb{E}}_{z\sim p_(z)}[D_{z,-1}(z)] + \mathop{\mathbb{E}}_{v\sim p_{emp}(v)}[{D_{z,-1}}(E(v))]+\lambda \mathop{\mathbb{E}}_{z\sim \hat{p}(z)}[(\nabla {D}_{z,-1}(z)-1)^2]\\
\end{aligned}
\right.
\end{equation}
where $p(z)$ and $p_{emp}(v)$ denote the multivariate Gaussian distribution and the empirical distribution of $\{\textbf{v}_i\}_{i=1}^n,$ respectively. $\hat{p}(z)$ and $\hat{p}(v)$ denoted the uniform sampling from the straight lines between the points sampled from observational data and generated data. To make the output of the binary classifier $D_z$ differentiable, we use $D_{z,-1}(z)$ to denote the output before binarization, where binarization is achieved by a sigmoid function. Minimizing the loss of a generator (e.g., $L_{GAN} (E)$) and the corresponding discriminator (e.g., $L_{GAN} (D_z)$) are adversarial as the two networks, $E$ and $D_z$, compete with each other during the training process. $\lambda$ is a penalty coefficient that is set to 10 in all experiments. The adversarial training of $G$ and $D_v$ is similar. 

In addition to the GAN-based adversarial training losses, we introduced reconstruction losses to ensure that the reconstructed data is closed to the real data. The losses are represented as 
$$\mathcal{L}_{rec}(E, G) = || z - E(G(z)) ||_2^2 + || v - G(E(v)) ||_2^2$$

Finally, to learn the generative models for the treatment and outcome variables, we impose the following mean squared error losses:
\begin{equation}
\left\{
\begin{aligned}
\mathcal{L}_{MSE}(F) =& || x - F(z_0,z_2) ||_2^2\\ \mathcal{L}_{MSE}(H) =&|| y - H(z_0,z_1,x) ||_2^2\\
\end{aligned}
\right.
\end{equation}

We summarize all the loss functions above and group them by discriminator networks $\mathcal{L}(D_z, D_v)$ and other networks $\mathcal{L}(G, E, F, H)$:

\begin{equation}
\left\{
\begin{aligned}
\mathcal{L}(G,E,F,H) =& \mathcal{L}_{GAN}(E)+\mathcal{L}_{rec}(E, G)+\mathcal{L}_{MSE}(F)+\mathcal{L}_{MSE}(H)\\ 
\mathcal{L}(D_z, D_v) =&\mathcal{L}_{GAN}(D_z)+\mathcal{L}_{GAN}(D_v)\\
\end{aligned}
\right.
\end{equation}
We alternatively update the parameters in one of $(E,G,F,H)$ or $(D_z,D_v)$ given the value of the other. The CausalEGM model is trained in an end-to-end fashion.

\subsection{Model architecture}
The architecture of CausalEGM is highly flexible. In this work, we use fully-connected layers for all networks. Specifically, the $(E, G, F, H)$ networks contain 5 fully-connected layers, and each layer has 64 hidden nodes. The $(D_z, D_v)$ networks each contain 3 fully-connected layers with 64, 32, and 8 hidden nodes, respectively. The leaky-ReLu activation function is deployed as a non-linear transformation in each hidden layer. We use Sigmoid as the activation function in the last layer of $H$ network when the treatment is binary. For continuous treatments, we do not use any activation function. Batch normalization \citep{pmlr-v37-ioffe15} is applied in discriminator networks. We use Adam optimizer \citep{adam2014} with initial learning rate as $2\times 10^{-4}$. The model parameters were updated in a mini-batch manner with the batch size equaling to 32. The default number of training iterations is 30,000.

\section{Theoretical Analysis}
We introduce a theoretical framework for analyzing GAN in Section~\ref{sec:GAN}. We then present the set up of our model in Section~\ref{sec:Problem_Setup}. Next, we discuss the excess risk bound for the CausalEGM model in Section~\ref{sec:Excess risk bound}. Finally, consistency analysis is provided based on an assumption related to the dimension reduction property for covariates in Section~\ref{sec: Consistency analysis}. Besides, we conducted experiments to demonstrate the rationality of the assumption.

\subsection{GAN Background} \label{sec:GAN}
Let $P$ and $Q$ be two probability measures and $\mathscr{A}$ be a class of measurable subsets of the space $\mathscr{X}$. Then define $d(P, Q;\mathscr{A}) \coloneqq \sup_{A\in \mathscr{A}}|P(A)-Q(A)|$ (Note if we let $\mathscr{B}$ be the Borel sets, $d(P,Q;\mathscr{B})$ would become the variation distance between $P$ and $Q$. Suppose $P$ and $Q$ have densities $p$ and $q$, we then have $d(P,Q;\mathscr{B})=\frac{1}{2}\lVert p-q \rVert_{L_{1}}=\frac{1}{2}\int |p-q|(x)d\mu(x)$.) Note that the function $d(\cdot)$ defines a pseudo-distance function between measures in the sense that it has the three following properties:

\begin{enumerate}[(i)]
\item $d(P, P;\mathscr{A})=0$
\item \textit{Symmetry}: $d(P, Q;\mathscr{A})=d(Q, P;\mathscr{A})$
\item \textit{Triangle inequality}: If $P'$ is another probability measure, we then have:
$$
d(P, Q;\mathscr{A}) \leq d(P, P';\mathscr{A}) + d(P', Q;\mathscr{A})
$$
\end{enumerate}

Let $\mathscr{A}_{M}\coloneqq \{A \in \mathscr{A}: \exists D\in \mathscr{D}_{M} \text{ s.t. } \forall x \in A, D(x)=1 \}$ where $D:\mathscr{X}\rightarrow \{0,1\}$ indicates a classifier and $\mathscr{D}_{M}$ is the set of classifiers constructed by deep neural networks with complexity parameter $M$ ($M$ can represent the number of layers, numbers of hidden nodes, etc). Let $P_{emp}=\frac{1}{n}\sum_{i=1}^{n}\delta_{x_i}(x)$ where $\{x_i:i=1,...,n\}$ is the set of observed samples. To train a classifier that distinguishes $Q$ and $P_{emp}$, we find $A^{\ast}$ s.t. $A^{\ast}=\arg\sup_{A\in{\mathscr{A}_{M}}}\lVert Q-P_{emp} \rVert_{L_{1}}$. WLOG, suppose $Q(A^\ast)-{P_{emp}}(A^\ast)>0$, then $\forall A \in \mathscr{A}_{M}$, $Q(A^\ast)-{P_{emp}}(A^\ast)\geq Q(A)-{P_{emp}}(A)$. Now suppose $A=I_{\{D=1\}}$ for some classifier $D$, the optimal discriminator $D^\ast$ between $Q$ and $P_{emp}$ would give $$Q(\{D^\ast(X)=1\})-{P_{emp}}(\{D^\ast(X)=1\})\geq Q(\{D(X)=1\})-{P_{emp}}(\{D(X)=1\})$$
for all $D$ in the class of discriminator under consideration.
Thus, if  $Q_G$ is the probability distribution for $G(Z)$, the adversarial training is then equivalent to minimizing the pseudo-distance between the induced distribution and the empirical distribution
$$\inf_G\sup_{A\in\mathscr{A}_{M}}|Q_{G}(A)-P_{emp}(A)|=\inf_{G}d(Q_G,P_{emp};\mathscr{A}_{M})$$
.

\subsection{Problem Setup and Notation} \label{sec:Problem_Setup}

Now for the CausalEGM, suppose
\begin{equation} \label{eqn: basic_eqn}
\left\{
\begin{aligned}
Y&= f^0(X,Z_0,Z_1)+\epsilon_1, \\
X&= h^0(Z_0,Z_2)+\epsilon_2,\\
(Z_0,Z_1,Z_2)&= (e_0^0(V),e_1^0(V),e_2^0(V)),\\
V&= g^0(Z),\\
\mathbb{E}[\epsilon_1] = \mathbb{E}[\epsilon_2]&= 0,\text{ }var(\epsilon_1) = \sigma_1,\text{ }var(\epsilon_2) = \sigma_2.
\end{aligned}
\right.
\end{equation}
where $f^0$, $h^0$, $e^0$, and $g^0$ are the unknown underlying functions that relate $Y$, $X$, $V$, and $Z$.
We aim to train $f$, $h$, $e$, and $g$ using CausalEGM to approximate $f^0$, $h^0$, $e^0$, and $g^0$. We want the latent variable $Z=(Z_0, Z_1, Z_2, Z_3)$ to have a fixed distribution (e.g., multivariate Gaussian) so that both $f$ and $h$ are generative models and $e$ is a good encoder that will capture most of the variation in V. Let $V\in \mathcal{V}$ be a continuous random variable in a $p$-dimensional space. We aim to learn two mappings $e:\mathcal{V}\rightarrow \mathbb{R}^q$ and $g:\mathbb{R}^q\rightarrow \mathcal{V}$ where $q \ll p$. Denote $Z^0$ as the random variable that follows a standard multivariate Gaussian distribution. Then we want our trained encoder $e$ to satisfy:
\begin{equation} \label{eqn:multinormal}
e(V)\sim Z^0
\end{equation}


In order for $g$ to reconstruct $e(V)$, $e(\cdot)$ and $g(\cdot)$ should minimize $\mathbb{E}_{0}||V-g(e(V))||_2^2$. 
Accordingly, we  design the loss functions:

\begin{equation}
\label{loss_funcs}
\left\{
\begin{aligned}
L_1=&\mathbb{E}_{n}||Y-f(X,e_0(V),e_1(V))||_2^2 \\
L_2=&\mathbb{E}_{n}||X-h(e_0(V),e_2(V))||_2^2 \\
L_3=&\sup_{A\in \mathscr{A}_m}|P(A;Z^0)-P_{emp}(A;e(V))|=d(P_{Z^0},P_{emp(e(V))};\mathscr{A}_m)\\
L_4=&\mathbb{E}_{n}||V-g(e(V))||_2^2
\end{aligned}
\right.
\end{equation}

where $\mathbb{E}_n$ is the empirical expectation. $P_{Z^0}$ is the probability measures of $Z^0$ and $P_{emp(e(V))}$ is the empirical distribution of $e(V)$. The empirical risk is denoted as follows:

$$R_{emp}=L_1+L_2+L_3+L_4$$ 

Hence the corresponding true risk is:
$$R^0=R^0_{1}+R^0_{2}+R^0_{3}+R^0_{4}$$
where
\begin{equation}
\label{true_risks}
\left\{
\begin{aligned}
R^0_{1}=&\mathbb{E}_{0}||Y-f(X,Z_0,Z_1)||_2^2 \\
R^0_{2}=&\mathbb{E}_{0}||X-h(Z_0,Z_2)||_2^2 \\
R^0_{3}=&d(P_{Z^0},P_{e(V)};\mathscr{A}_m)\\
R^0_{4}=&\mathbb{E}_{0}||V-g(e(V))||_2^2
\end{aligned}
\right.
\end{equation}
Note that $\mathbb{E}_0$ stands for the expectation w.r.t the underlying distribution of the random variables and $P_{e(V)}$ is the probability measure induced by $e(V)$.
We denote $\mathscr{F}_M$ as the class of deep neural networks of complexity $M$. Let $\hat{f}_M, \hat{h}_M, \hat{e}_M$ and $\hat{g}_M$ to be the solution of $\inf_{f,h,e,g\in \mathscr{F}_M}R_{emp}(f,h,e,g).$ Let $f^0_M, h^0_M, e^0_M$ and $g^0_M$ to be the solution of $\inf_{f,h,e,g\in \mathscr{F}_M}R^0(f,h,e,g)$. So $(\hat{f}_M,\hat{h}_M, \hat{e}_M,\hat{g}_M)$ is the solution that minimizes empirical risk and $(f^0_M, h^0_M, e^0_M, g^0_M)$ is the solution that minimizes the true risk. 

\subsection{Excess risk bound}\label{sec:Excess risk bound}
We can now define the excess risk within the class $\mathscr{F}_M$: 
$$
R^0(\hat{f}_M,\hat{h}_M, \hat{e}_M,\hat{g}_M)-\inf_{f,h,e,g\in \mathscr{F}_M}R^0(f,h,e,g) = R^0(\hat{f}_M,\hat{h}_M, \hat{e}_M,\hat{g}_M)-R^0(f^0_M, h^0_M, e^0_M, g^0_M)
$$

Before moving forward, we first make some assumptions on the $\mathscr{F}_M$.

\begin{assumption}\label{as:uniformly_bounded}
(\textbf{$b$-uniformly bounded})
$\bigcup_{M}\mathscr{F}_{M}$ is $b$-uniformly bounded:  there exists some $b>0$ such that for any $f\in \bigcup_{M}\mathscr{F}_{M}$, $||f||_{\infty}\leq b.$ 
\end{assumption}

\begin{assumption}\label{as:uniformly_equi-continuous}
(\textbf{uniformly equi-continuous})
$\bigcup_{M}\mathscr{F}_{M}$ is uniformly equi-continuous:
$\forall \epsilon>0$, there exists a $\delta>0$ such that for any $f_1, f_2 \in \bigcup_{M}\mathscr{F}_{M}$,
$$
|f_1(x)-f_2(y)|<\epsilon
$$
whenever $|x-y|<\delta$.
\end{assumption}

Note the above assumptions can be ensured by putting constraints on the gradients of functions and bounding the domain of the input space. We are now able to show that the excess risk converges to zero in probability as  $M,n\rightarrow\infty$ where $M$ denotes the complexity of the class of neural networks and $n$ denotes the number of sample points. Denote $P_{Z_{emp}}$ as the empirical distribution of $Z$.

\begin{lemma}

$$R^0(\hat{f}_M,\hat{h}_M, \hat{e}_M,\hat{g}_M) \leq \inf_{f,h,e,g\in \mathscr{F}_M}R^0(f,h,e,g) +\alpha_{M,n}+\beta_{M,n}+\gamma_{M,n}+\zeta_{M,n}$$

where 
\begin{equation}
\left\{
\begin{aligned}
\alpha_{M,n}=&2\sup_{f\in \mathscr{F}_M}|(\mathbb{E}_n-\mathbb{E}_0)(||Y-f(X,Z_0,Z_1)||_2^2)|\\ \beta_{M,n}=&2\sup_{h\in \mathscr{F}_M}|(\mathbb{E}_n-\mathbb{E}_0)(||X-h(Z_0,Z_2)||_2^2)|\\ 
\gamma_{M,n}=&2d(P_{Z_{emp}},P_{Z^0};\mathscr{A}_M)\\ \zeta_{M,n}=&2\sup_{g,e\in \mathscr{F}_M}|(\mathbb{E}_n-\mathbb{E}_0)(||V-g(e(V))||_2^2)|
\end{aligned}
\right.
\end{equation}
\label{thm:estimation_error_inequality}
\end{lemma}

\begin{theorem} \label{thm: Estimation_Error}
(\textbf{Bound of Excess Risk})
Denote $O:= (Y, X, V, Z_0, Z_1, Z_2).$ Then we define the Rademacher complexity of a real-valued function class $\mathscr{F}$ as
$\mathscr{R}_n(\mathscr{F}):=\mathbb{E}_{\epsilon,O}[\sup_{f\in\mathscr{F}}|\frac{1}{n}\sum_{i=1}^{n}\epsilon_if(O_i)|]$ where $\epsilon_{1},\epsilon _{2},\dots ,\epsilon _{n}$ are independent random variables drawn from the Rademacher distribution. 
For any $\delta>0$, we have

$$R^0(\hat{f}_M,\hat{h}_M, \hat{e}_M,\hat{g}_M)-\inf_{f,h,e,g\in \mathscr{F}_M}R^0(f,h,e,g)\leq12\mathscr{R}_n(\mathscr{F}_M)+ 4\mathscr{R}_n(\mathscr{D}_{M})+\delta$$
with probability at least $1-4e^{-\frac{n\delta^2}{512b^2}}$.
\end{theorem}

There are several existing works that provide an upper bound to the Rademacher complexity of neural networks. For example, when the set of functions in $\mathscr{F}$ is 1-Lipschitz, ~\cite{bartlett2017spectrally} obtained a upper bound of $\mathcal{O}\left(b\sqrt{\frac{D^3}{n}}\right)$ to $\mathscr{F}$ where $D$ denotes the number of layers of network $f \in \mathscr{F}$. Similarly, ~\cite{li2018tighter} gave a bound of $\mathcal{O}\left(b\sqrt{\frac{Dk^2}{n}}\right)$, where $k$ denotes the dimension of features.




\subsection{Consistency analysis}\label{sec: Consistency analysis}

With one more assumption, we can then show the consistency of our neural networks.


\begin{assumption}\label{as:reconstruction_error}
There exists $\tilde{e}_3$, $\tilde{g}$ and $\delta>0$ s.t.

\begin{equation} \label{eqn:latent distribution}
(e_0^0, e_1^0, e_2^0, \tilde{e}_3) \overset{\mathcal{D}}{=}Z^0
\end{equation}
where the quadruplet denotes the four components of the encoder function and

for any function $e$ and $g$,
\begin{equation} 
\mathbb{E}_{0}||V-\tilde{g}((e_0^0, e_1^0, e_2^0, \tilde{e}_3)(V))||_2^2\leq\mathbb{E}_{0}||V-g((e)(V))||_2^2 + \delta
\end{equation}
\end{assumption}



This assumption is expected to hold with a small delta when the distribution of V satisfies a certain "dimension reduction" property. We provide a concrete simulation example in appendix~\ref{sim_asm4} to demonstrate this.

 From Assumption~\ref{as:uniformly_bounded} and Assumption~\ref{as:uniformly_equi-continuous} we conclude that $\bigcup_{M}\mathscr{F}_{M}$ is sequentially compact by the \emph{Arzelà–Ascoli} theorem. Hence $\bigcup_{M}\mathscr{F}_{M}$ is a \emph{Glivenko-Cantelli} class and $\mathscr{R}_n(\bigcup_{M}\mathscr{F}_{M}) \rightarrow 0$ as $n \rightarrow \infty$. We also have $\mathscr{R}_n(\mathscr{D}_{M}) \rightarrow 0$ as $\mathscr{D}_{M}$ has finite VC-dimension. Now let $m:\mathbb{N}\rightarrow\mathbb{N}$ be a strictly increasing function s.t. $\lim_{n\rightarrow\infty}\mathscr{R}_n(\mathscr{F}_{m_n}) = 0$ and $\lim_{n\rightarrow\infty}\mathscr{R}_n(\mathscr{D}_{m_n}) = 0$.  Let $(\hat{f}, \hat{h}, \hat{e}, \hat{d})_n$ be the sequence of quadruple that solves $\inf_{f,h,e,g\in \mathscr{F}_{m_n}}R_{emp}$, then there exists a limit point of this sequence as $n \rightarrow \infty$.

\begin{theorem} \label{thm: consistency}
(\textbf{Consistency})

Suppose assumptions $1$-$4$ hold. Let $(f^{\ast},h^{\ast},e^{\ast},g^{\ast})$ be any limit point of $(\hat{f},\hat{h},\hat{e},\hat{g})_n$. We then have

$$
\mathbb{E}_{0}||(f^0-f^{\ast})(X,Z_0,Z_1)||_2^2+\mathbb{E}_{0}||(h^0-h^{\ast})(Z_0,Z_2)||_2^2+d(P_{Z^0},P_{e^{\ast}(V)};\mathscr{A}_M)\\ \leq 2\delta.
$$

\end{theorem}

Theorem~\ref{thm: consistency} suggests that if $V$ can be encoded effectively s.t. the Assumption~\ref{as:reconstruction_error} are satisfied with $\delta \approx 0$, we would have approximately

$$
f^{\ast}\approx f^0, h^{\ast}\approx h^0, e^{\ast}(V) \overset{\mathcal{D}}{\approx}Z^0.
$$
This holds for any limit points of $\{(\hat{f},\hat{h},\hat{e},\hat{g})_n\}$.

\section{Experiments}
\label{sec: Experiments}
We performed a series of experiments to evaluate the performance of CausalEGM against some state-of-the-art methods. In observational studies, accurately estimating the treatment effects on the population level and individual level are both crucial.  We aim to verify the ability of CausalEGM to estimate both the average treatment effect on the population level and the individual treatment estimation concerning the heterogeneous treatment effects. Since CausalEGM is applicable for both binary treatment and continuous treatment, we test the performance of CausalEGM under both settings. 

\subsection{Datasets}
For the continuous treatment setting, three simulation datasets and one real dataset from previous publications will be used. 
\paragraph{Hirano and Imbens}
We follow a similar data-generating process as in ~\cite{hirano2004propensity} and ~\cite{moodie2012estimation} as follows:
let $V_1$, $V_2$,...,$V_{p}$ be i.i.d. unit exponential random variables, $Z_0 = V_{1}$, $Z_1 = V_{2}$, $Z_2 = V_{3}$, $X|V \sim exp(Z_0 + Z_1)$, and $Y(x)|V \sim N(x + (Z_0 + Z_2)exp(-x(Z_0 + Z_2), 1)$. Then the dose-response function can be obtained by integration w.r.t the covariates $V$: $\mu(x) = x + \frac{2}{(1 + x)^3}.$ We use $p=200$ in the simulation experiment.

\paragraph{Sun}
We generate a synthetic dataset using a similar data generating process described in ~\cite{sun2015causal}. with some modifications to fit for continuous treatment.
Specifically, we let  $V_1,...,V_p\stackrel{iid}\sim N(0,1)$ and define $f_1(u)=-2sin(2u)$, $f_2(u)=u^2-\frac{1}{3}$, $f_3(u)=u-\frac{1}{2}$, $f_4(u)=cos(u)$, $f_5(u)=u^2$ and $f_6(u)=u$. We then generate the treatment to be $X\sim N(\sum_{i=1}^4f_{i}(V_i), 1)$  and the outcome to be $Y\sim N(X+f_{3}(V_1)+f_{4}(V_2)+f_{5}(V_5)+f_{6}(V_6), 1)$. Then the dose-response function can be obtained by integration w.r.t the covariates $V$: $\mu(x) = x + 0.5+e^{-0.5}.$ We use $p=200$ in the simulation experiment.

\paragraph{Colangelo and Lee}
We followed a similar data generation process in ~\cite{colangelo2020double} as follows: let $\epsilon_1\sim N(0,1)$, $\epsilon_2\sim N(0,1)$. The covariates are generated by $V=(V_1,...,V_{p})^{'} \sim N(0,\Sigma)$ where $diag(\Sigma)=1$ and $\Sigma_{i,j}=0.5$ for $|i-j|=1$. The treatment is generated by $X = \Phi(3V^{'}\theta)+0.75\epsilon_1-0.5$ where $\theta_j=1/j^2$. The outcome is generated by $Y=1.2X+1.2V^{'}\theta+X^{3}+XV_{1}+\epsilon_2$. The dose-response function is $\mu(x) =1.2x+x^{3}.$ We use $p=200$ in the simulation experiment.

\paragraph{Twins}
This dataset contains data of 71,345 twins, including their weights (used as treatment), mortality, and 50 other covariates (so $p=50$) derived from all births in the USA between 1989-1991. Similar to ~\cite{pmlr-v127-li20a}, we first filtered the data by limiting the weight to be less than 2 kilograms. 4821 pairs of twins were kept for further analysis.  We set the weights as the continuous treatment variable. We then simulate the risk of death (outcome) under a model in which higher weight leads to a lower death rate in general. Let $Y$ be the Bernoulli variable where $Y=1$ indicates death and $R$ be the death risk that depends on the covariates. We simulate the outcome as $Y(x)\sim Bernoulli(R(x)),$  $R(x) = -\frac{2}{1+e^{-3X}} + v\gamma + \epsilon$
where $\gamma \in \mathbb{R}^{p\times1}$ and $\gamma_i \sim N(0, 0.025^2)$, $\epsilon \sim N(0, 0.25^2)$. Taking the response variable as $R$ instead of $Y,$ the ADRF is then: $\mu(x) = -\frac{2}{1+e^{-3x}} + \mathbb{E}[V_i \cdot \gamma]$.

\paragraph{ACIC 2018}
For binary treatment settings, we downloaded the datasets from the 2018 Atlantic Causal Inference Conference (ACIC) competition. This dataset utilizes the Linked Births and Infant Deaths Database (LBIDD) based on real-world medical measurements collected from \cite{ibm_data}. The LBIDD data is semi-synthetic where 117 measured covariates are given, and the treatment and outcome are simulated based on different data-generating processes. We chose nine datasets by selecting the most complicated generation process (e.g., the highest degree of generation function) with sample size ranging from 1,000 to 50,000. The details for the datasets used in the study are provided in appendix \ref{app_data}.

\subsection{Evaluation metrics}
In the continuous treatment setting, we aim to evaluate whether the estimated dose-response function $\mu(x)$ can well approximate the true dose-response function. Three different metrics are used.
\paragraph{Root-Mean-Square Error (RMSE)}
\begin{equation}
RMSE=\sqrt{\frac{1}{n}\sum_{i=1}^n||\mu(x_i)-\hat{\mu}(x_i)||_2^2} 
\end{equation}

\paragraph{Mean Absolute Percentage Error (MAPE)}
\begin{equation}
MAPE=\frac{1}{n}\sum_{i=1}^n||\frac{\mu(x_i)-\hat{\mu}(x_i)}{\mu(x_i)}||_1 
\end{equation}

\paragraph{Mean Absolute Error of MTFE (Bias(MTFE))}
We first calculate \textit{Marginal Treatment Effect Function} (MTFE) at $x$ as:
\begin{equation}
MTEF=\frac{\mu(x+\Delta x)-\mu(x)}{\Delta x}
\end{equation}
Then Mean Absolute Error of MTFE (Bias(MTFE) is then defined to be the mean absolute difference between the ground truth MTFE using $\mu(x)$ and the estimated MTFE using $\hat{\mu}(x)$. In practice, we choose $\Delta x$ to be $0.0001$.

In the binary treatment settings, we consider both the population-level average treatment effect and individual treatment effect. Two commonly used metrics are introduced as follows.

\paragraph{Absolute Error in Average Treatment Effect $(\epsilon_{ATE})$}
$$
\epsilon_{ATE} = |\frac{1}{n}\sum_{i=1}^{n}{}{(\hat{Y}_i(1)- \hat{Y}_i(0))}-\frac{1}{n}\sum_{i=1}^{n}{}{(Y_i(1)- Y_i(0))}|
$$

\paragraph{Precision in Estimation of Heterogeneous Effect $(\epsilon_{PEHE})$}
$$
\epsilon_{PEHE}=\frac{1}{n}\sum_{i=1}^{n}(\hat{Y}_i(1)-\hat{Y}_i(0)-(Y_i(1)-Y_i(0)))^2
$$
where $\hat{Y}_i(\cdot)$ denotes the predicted/imputed value of potential outcome.

\subsection{Baselines}
For continuous treatment setting, three different baselines were used. 

\textbf{Ordinary Least Squares regression (OLS)}. OLS first fit a  linear regression model for $Y|(X,V)$. For each value of treatment $x$, the estimated ADRF is then $\frac{1}{n}\sum_{i}^{n}ls(x, {v}_{i})$ where $ls$ is the fitted linear model.

\textbf{Regression Prediction Estimator (REG)}~\citep{schafer2015causal, galagate2016causal, imai2004causal}. The \textit{prima facie}  estimator is an estimator that regresses the outcome on the treatment without considering covariates. REG generalizes the notion of \textit{prima facie} estimator. It takes the covariates into account when doing regression. Unlike OLS, REG fits a quadratic ADRF: $Y(x) = \alpha_0 + \alpha_1 x + \alpha_2 x^2.$

\textbf{Double Debiased Machine Learning Estimator (DML)}. See ~\cite{colangelo2020double}. DML is a kernel-based machine learning approach that combines a doubly moment function and cross-fitting. Various machine learning methods can be used to estimate the conditional expectation function and conditional density. We used "Lasso", "random forest", and "neural network" provided by the DML toolkit as three variants, denoted as DML(lasso), DML(rf) and DML(nn).

For the binary treatment setting, five baselines were introduced.

\textbf{CFR}. CFR \cite{shalit2017estimating} estimated the individual treatment effect (ITE) by utilizing neural networks to learn the low-dimensional representation for covariates and two outcome functions, respectively. An integral probability metric was further introduced to control the balance of distributions in the treated and control group. We use the two variants of CFR for comparison, which are referred to as TAENET and CFRNET.

\textbf{Dragonnet}. \cite{shi2019adapting} used a three-head architecture, which contains a two-head architecture for outcome estimation and a one-head architecture for propensity score estimation. It is noted that Dragonnet uses essentially the same architecture as CFR if the propensity-score head is removed.

\textbf{CEVAE}. \cite{louizos2017causal} is a variational autoencoder-based method for estimating the treatment effect where the above CFR architecture \citep{shalit2017estimating} was used in the inference network and the latent variables were set to be multivariate normal distribution. 

\textbf{GANITE}. \cite{yoon2018ganite} exploited a generative adversarial network (GAN) model for generating the counterfactual outcome through adversarial training. 

\textbf{Causalforest}.\cite{wager2018estimation} built random forests to estimate the heterogeneous treatment effect that is applicable in binary treatment settings. Causalforest is an ensemble method that consists of multiple causal trees.

\subsection{Results}
\begin{figure}[htbp]
  \centering
  \includegraphics[width=1.0\columnwidth]{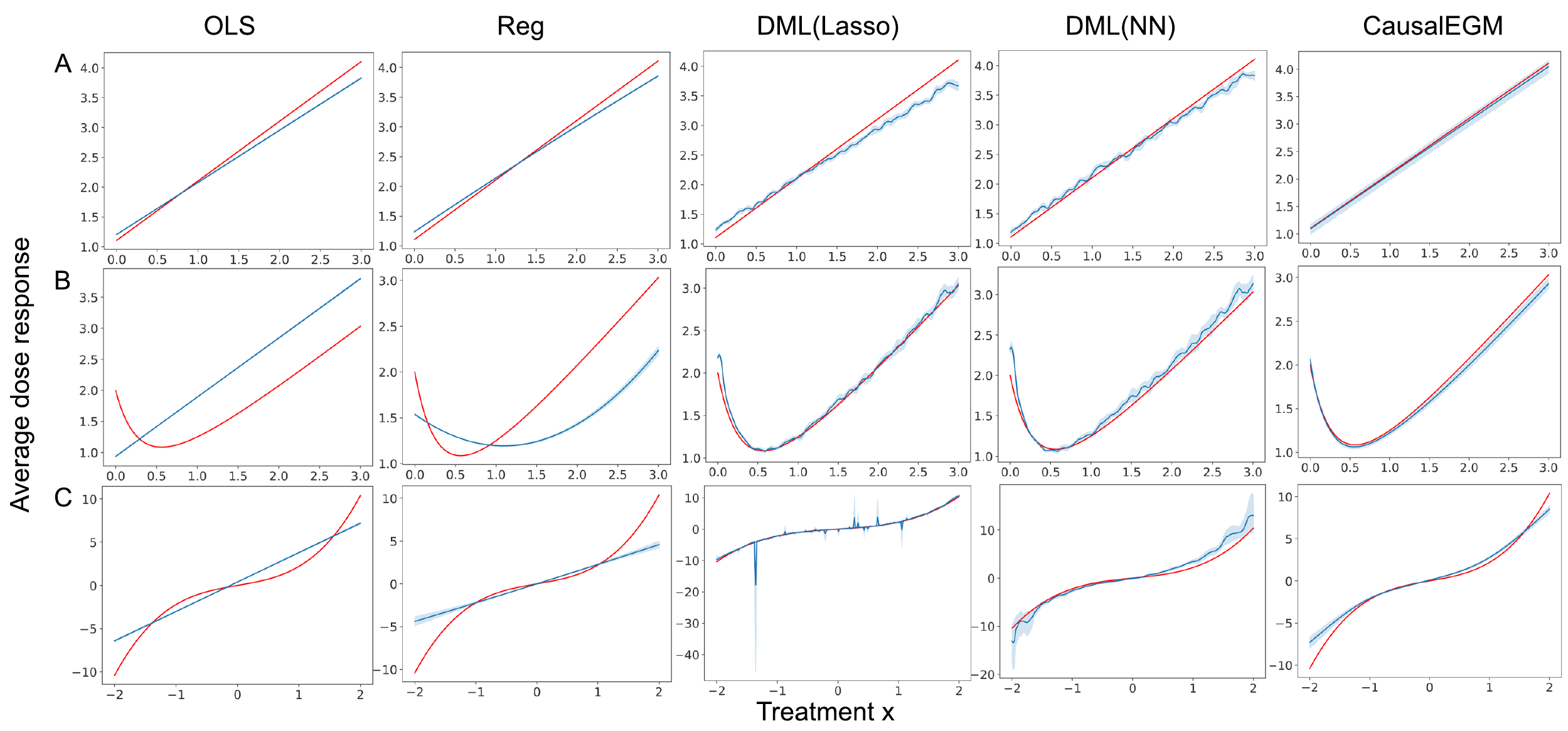}
  \caption{The performance of CausalEGM and baseline methods (OLS, Reg, DML with Lasso or neural network) under continuous treatment settings across three benchmark datasets. (A) Hiranos and Imbens dataset. (B) Sun et al dataset. (C) Colangelo and Lee dataset. The red curves are the ground truth while the blue curves are the estimated average dose-response with 95\% confidence interval based on 10 independent simulations.}
  \label{fig: continuous result}
\end{figure}

We first evaluate the performance of CausalEGM model in the continuous setting where the treatment $x\in \mathscr{X}$ and $\mathscr{X}$ is a bounded interval in $\mathbb{R}$. We compare CausalEGM with four different methods, including one using neural networks as its key component. It is shown that CausalEGM demonstrates superior results over the existing methods, including two linear regression-based methods OLS and REG, and a kernel-based machine learning approach with two different machine learning algorithms (lasso and neural network). We first evaluate whether the dose-response function can be well estimated by different competing methods (see Figure \ref{fig: continuous result}). It is observed that OLS and Reg result in relatively large estimation errors. The dose-response curves estimated by the DML methods have spikes and fluctuation. In contrast, the curves estimated by CausalEGM are smooth and the estimation errors are small.

In terms of the quantitative measurements, CausalEGM achieves the lowest RMSE, MAPE, Bias(MTEF) in all three simulation datasets compared to baseline methods (Table \ref{table:continuous}). We also note that DML method performs much better than linear regression-based methods (OLS and REG) in Hiranos and Imbens and Twins datasets while performing less well in the other two. CausalEGM reduces the RMSE, MAPE, Bias(MTEF) by 24.2\% to 63.4\%, 6.9\% to 55.2\%, and 17.8\% to 84.6\% compared to the best baseline method across different datasets, respectively. The results of both simulation data and real data illustrate that CausalEGM offers significant improvement for estimating the causal effect in continuous settings.

 \renewcommand\arraystretch{0.6}
\begin{table}
  \centering
  \resizebox{\textwidth}{!}{
  \begin{tabular}{cccccl}
    \toprule
    Dataset&Method&RMSE&MAPE&Bias(MTEF) \\
    \midrule
    \multirow{6}*{Hiranos and Imbens}  
    &OLS&$0.680\pm0.0$&$0.367\pm0.0$&$0.629\pm0.0$\\
    &REG&$0.525\pm0.0$&$0.214\pm0.0$&$0.586\pm0.0$\\
    &DML(lasso)&$0.090\pm0.0$&$0.037\pm0.0$&$1.8\pm0.0$\\
    &DML(nn)&$0.133\pm0.022$&$0.052\pm0.011$&$653\pm406$\\   &\textbf{CausalEGM}&$\bm{0.041\pm0.014}$&$\bm{0.019\pm0.006}$&$\bm{0.082\pm0.017}$\\
    \midrule
    \multirow{6}*{Sun et al}     
    &OLS&$0.140\pm0.0$&$0.041\pm0.0$&$0.124\pm0.0$\\
    &REG&$0.117\pm0.0$&$0.039\pm0.0$&$0.127\pm0.0$\\
    &DML(lasso)&$0.163\pm0.0$&$0.050\pm0.0$&$0.719\pm0.0$\\
    &DML(nn)&$0.0970\pm0.0190$&$0.0346\pm0.006$&$300\pm249$\\
    
    &\textbf{CausalEGM}&$\bm{0.0738\pm0.0399}$&$\bm{0.0345\pm0.0170}$&$\bm{0.0247\pm0.0148}$\\
    \midrule
    \multirow{6}*{Colangelo and Lee} &OLS&$1.3\pm0.0$&$1.2\pm0.0$&$2.0\pm0.0$\\
    &REG&$1.5\pm0.0$&$0.565\pm0.0$&$2.0\pm0.0$\\
    &DML(lasso)&$0.487\pm0.0$&$0.168\pm0.0$&$1.8\pm0.0$\\
    &DML(nn)&$1.3\pm0.581$&$0.494\pm0.181$&$3600\pm4400$\\
    
    &\textbf{CausalEGM}&$\bm{0.125\pm0.040}$&$\bm{0.119\pm0.080}$&$\bm{0.216\pm0.048}$\\
    \midrule
    \multirow{6}*{Twins} &OLS&$0.109\pm0.0$&$0.260\pm0.0$&$0.319\pm0.0$\\
    &REG&$11\pm0.0$&$64\pm0.0$&$2.1\pm0.0$\\
    &DML(lasso)&$0.075\pm0.0$&$0.165\pm0.0$&$7.7\pm0.0$\\
    &DML(nn)&$0.059\pm0.002$&$0.158\pm0.006$&$56\pm15$\\
    
    &\textbf{CausalEGM}&$\bm{0.0339\pm0.020}$&$\bm{0.090\pm0.053}$&$\bm{0.491\pm0.021}$\\
  \bottomrule
\end{tabular}}
\caption{Result on synthetic datasets with sample size $n=20000$, number of covariates $p=200$. Each method was run for 10 times and the standard deviation was also shown.}
\label{table:continuous}
\end{table}
In the binary treatment settings where the treatment $x\in \{0,1\}$, we aim to evaluate whether CausalEGM could estimate an accurate treatment effect. CausalEGM was benchmarked against a number of state-of-the-art methods on the LBIDD benchmark datasets, which provide various simulation settings and sample sizes. We chose three datasets from each of three different sample sizes (1k, 10k, and 50k) with the most complicated generation process (e.g., the generation functions are of the highest order/degree). CausalEGM is compared to six baseline methods on each of these datasets.
As shown in Table \ref{table:binary}, CausalEGM achieves the smallest $\epsilon_{ATE}$ in 6 out of 9 datasets. CausalEGM performs especially well in datasets with large sample sizes (e.g., 50k). For example, the $\epsilon_{ATE}$ is reduced by 16.7\% to 98.7\% in the three largest datasets compared to the second-best method. For another metric, CausalEGM achieves the smallest $\epsilon_{PEHE}$ in 5 out of 9 datasets and the second-best performance in the rest 4 datasets. To sum up, our model shows superior performance in estimating both average treatment effect and individual treatment effect and is substantially powerful when the sample size is large.

\renewcommand\arraystretch{1.2}
\begin{table}
  \centering
  \resizebox{\textwidth}{!}{
  \begin{tabular}{ccccccccc}
    \toprule
    Metric&Dataset & TARNET & CFRNET & CEVAE & GANITE & Dragonnet&CausalForest& \textbf{CausalEGM}\\
    \midrule
    \multirow{9}*{$\epsilon_{ATE}$}&\multirow{3}*{Datasets-1k} & $0.022\pm0.015$ & $0.018\pm0.015$ & $0.035\pm0.021$ &$0.27\pm0.08$& $0.010\pm0.004$& $0.021\pm 0.001$&$\bm{0.0097\pm0.0075}$\\
    & & $0.038\pm0.029$&$0.041\pm0.027$&$0.12\pm0.10$&$2.0\pm0.3$&$\bm{0.012\pm0.007}$&$0.017\pm 0.003$&$0.032\pm0.020$\\
    & & $0.10\pm0.06$&$\bm{0.095\pm0.079}$&$0.38\pm0.27$&$2.0\pm1.4$&$0.16\pm0.10$&$0.23\pm0.02$&$0.26\pm0.07$\\  
    \cline{2-9}
    &\multirow{3}*{Datasets-10k}&$6.4\pm3.5$&$12\pm7$&$204\pm58$&${2.7\pm1.2}$&$124\pm11$&$2.5\pm 1.1$&$\bm{1.3\pm0.6}$\\
    & & $0.056\pm0.001$&$0.056\pm0.001$&$0.070\pm0.031$&$1.2\pm0.2$&$0.0097\pm0.069$&$0.0057\pm 0.0004$&$\bm{0.0043\pm0.0025}$\\
    & & $0.034\pm0.023$&$0.060\pm0.002$&$0.018\pm0.011$&$0.12\pm0.09$&$0.078\pm0.057$&$\bm{0.013\pm 0.003}$&${0.039\pm0.016}$\\
    \cline{2-9}
    &\multirow{3}*{Datasets-50k}&$0.038\pm0.021$&$0.085\pm0.105$&$0.59\pm0.31$&$1.4\pm0.5$&$0.89\pm0.53$&$0.024\pm 0.003$&$\bm{0.020\pm0.013}$\\
    &&$0.044\pm0.003$&$0.045\pm0.004$&$0.66\pm0.59$&$2.3\pm0.2$&$0.027\pm0.028$&$0.010\pm 0.001$&$\bm{0.0098\pm0.0089}$\\
    &&$0.30\pm0.01$&$0.30\pm0.01$&$0.64\pm0.45$&$1.9\pm0.3$&$0.16\pm0.08$&$0.12\pm 0.01$&$\bm{0.0016\pm0.0010}$\\
    \midrule
    \multirow{9}*{$\epsilon_{PEHE}$}&\multirow{3}*{Datasets-1k} & $0.11\pm0.02$ & $\bm{0.00069\pm0.00075}$ & $0.012\pm0.005$ &$0.14\pm0.04$& ${0.038\pm0.003}$&$0.00080\pm 0.00005$&$0.0069\pm0.0016$\\
    & & $0.35\pm0.03$&$0.29\pm0.04$&${0.27\pm0.04}$&$4.34\pm1.24$&$0.34\pm0.01$&$0.27\pm 0.01$&$\bm{0.25\pm0.01}$\\
    & & $0.31\pm0.14$&$0.28\pm0.23$&$7.6\pm5.3$&$12\pm6$&$1.7\pm0.4$&$\bm{0.075\pm 0.006}$&${0.20\pm0.03}$\\
    \cline{2-9}
    &\multirow{3}*{Datasets-10k}&$433\pm106$&$662\pm288$&$46200\pm15500$&$78.7\pm26.8$&$22200\pm4130$&$483.72\pm 31.68$&$\bm{7.2\pm2.6}$\\
    & & $0.024\pm0.005$&$0.022\pm0.006$&$0.091\pm0.019$&$2.08\pm0.45$&$0.042\pm0.003$&${0.015\pm 0.001}$&$\bm{0.014\pm0.001}$\\
    & & $0.012\pm0.005$&$0.0040\pm0.0028$&$0.0034\pm0.0013$&$0.14\pm0.08$&$0.036\pm0.015$&$\bm{0.0016\pm 0.0008}$&${0.0028\pm0.0013}$\\
    \cline{2-9}
    &\multirow{3}*{Datasets-50k}&$0.88\pm0.04$&$0.90\pm0.08$&$1.1\pm0.5$&$3.4\pm1.4$&$1.84\pm0.83$&${0.65\pm 0.01}$&$\bm{0.55\pm0.01}$\\
    &&$0.031\pm0.006$&$0.030\pm0.011$&$0.84\pm0.76$&$5.454\pm0.65$&$0.039\pm0.007$&$\bm{0.020\pm 0.002}$&${0.022\pm0.001}$\\
    &&$0.22\pm0.07$&$0.27\pm0.05$&$0.67\pm0.61$&$3.8\pm1.1$&$0.14\pm0.06$&$0.022\pm 0.001$&$\bm{0.0054\pm0.0013}$\\

  \bottomrule
\end{tabular}}
 
\caption{The performance of CausalEGM and comparison methods in ACIC 2018 dataset with various sample sizes. Each method was run 10 times and the standard deviations are shown. The best performance is marked in bold.}
\label{table:binary}
\end{table}

We have demonstrated that the performances of CausalEGM under both continuous and binary treatment settings are superior. Since our model is composed of multiple neural networks, it is of interest to evaluate the contribution of different components. We first evaluated the contribution of the Roundtrip module. To do this, we removed the $G$ network and the discriminator networks $D_z$ and $D_v$ and denoted the model as CausalEGM w/o RT, which no longer requires adversarial training and reconstruction for $v$ and $z$. Taking the continuous treatment setting for an example, we note that the performance of CausalEGM without the Roundtrip module has a noticeable decline in all datasets. The RMSE, MAPE, Bias(MTEF) increases by 32.68\% to 164.88\%, 14.55\% to 376.95\%, and 21.85\% to 43.98\%, respectively (Table \ref{table:roundtrip}). Such experimental results imply that the adversarial training and the reconstruction error are essential for learning a good low-dimensional representation of the high-dimensional covariates.

\renewcommand\arraystretch{0.5}
\begin{table}
  \centering
  \resizebox{\textwidth}{!}{
  \begin{tabular}{cccccl}
    \toprule
    Dataset&Method&RMSE&MAPE&Bias(MTEF) \\
    \midrule
    \multirow{2}*{Hiranos and Imbens}  
    &CausalEGM w/o RT&$0.0936\pm0.0579$&$0.0434\pm0.0293$&$0.128\pm0.0624$\\
    &CausalEGM&$\bm{0.0706\pm0.0445}$&$\bm{0.0352\pm0.0210}$&$\bm{0.0889\pm0.0210}$\\
    \midrule
    \multirow{2}*{Sun et al}     
    &CausalEGM w/o RT&$0.106\pm0.0473$&$0.0438\pm0.0224$&$0.0305\pm0.0147$\\
    &CausalEGM&$\bm{0.0436\pm0.0085}$&$\bm{0.0180\pm0.0038}$&$\bm{0.0230\pm0.0116}$\\
    \midrule
    \multirow{2}*{Colangelo and Lee}
    &CausalEGM w/o RT&$1.28\pm0.129$&$0.488\pm0.0950$&$2.21\pm0.109$\\
     &CausalEGM&$\bm{0.886\pm0.232}$&$\bm{0.426\pm0.124}$&$\bm{1.66\pm0.290}$\\
    \midrule
    \multirow{2}*{Twins} 
    &CausalEGM w/o RT&$0.0641\pm0.0252$&$2.38\pm6.64$&$0.0903\pm0.0346$\\
     &CausalEGM&$\bm{0.0242\pm0.0132}$&$\bm{0.499\pm1.39}$&$\bm{0.0746\pm0.0314}$\\
  \bottomrule
\end{tabular}}
 \caption{Ablation study on evaluating the contribution of Roundtrip module. Each method was run for 10 times and the standard deviations are shown.}
\label{table:roundtrip}
\end{table}

Next, we investigate whether the adversarial training and the reconstruction error objectives are necessary in the bi-directional transoformation module. In our model design, adversarial training in latent space is necessary to guarantee the independence of latent variables. The reconstruction of $V$ is also required for ensuring the latent features contain all the information possessed by the original covariates. So we designed experiments to quantitatively evaluate the contribution of the adversarial training in covariate space and the reconstruction in latent space. As shown in Table \ref{table:ablation}, the reconstruction of latent features could help benefit the model training and achieve slightly better performance. Using the adversarial training in covariate space might not improve the model training as the distribution matching in high-dimensional space might be difficult.

Finally, we conduct comprehensive experiments to examine the robustness and scalability of CausalEGM (appendix \ref{app_scala}). Specifically, we first verify whether CausalEGM is sensitive to the choice of latent feature dimensions, which includes the total dimension of latent space and also the dimension of $Z_0$. The experimental results show that CausalEGM is quite robust to the choice of latent feature dimensions. For the scalability test, we demonstrate that CausalEGM is capable of handling datasets with a large number of covariates and large sample size (e.g., millions of data points).

\begin{table}
  \centering

  \resizebox{\textwidth}{!}{
  \begin{tabular}{cccccl}
    \toprule
    Dataset&(V-GAN, Z-Rec)&RMSE&MAPE&Bias(MTFE) \\

    \midrule
    \multirow{4}*{Hiranos and Imbens}  
    &(1,1)&$0.0906\pm0.0270$&$0.0439\pm0.0116$&$0.104\pm0.200$\\
    &(0,1)&$\bm{0.0727\pm0.0451}$&$\bm{0.0345\pm0.0190}$&$\bm{0.0890\pm0.0230}$\\
    &(1,0)&$0.0845\pm0.0321$&$0.0401\pm0.00984$&$0.0940\pm0.0355$\\
    &(0,0)&$0.0784\pm0.0363$&$0.0371\pm0.0163$&$0.103\pm0.0352$\\
                        
    \midrule
    \multirow{4}*{Sun et al}                 
    &(1,1)&$0.0567\pm0.0299$&$0.0219\pm0.0134$&$0.0280\pm0.0282$\\
    &(0,1)&$\bm{0.0436\pm0.00857}$&$\bm{0.0180\pm0.00388}$&$\bm{0.0230\pm0.0116}$\\
    &(1,0)&$0.0592\pm0.0202$&$0.0227\pm0.00826$&$0.03002\pm0.0267$\\
    &(0,0)&$0.0622\pm0.0322$&$0.0234\pm0.0124$&$0.0266\pm0.0252$\\
  \bottomrule
\end{tabular}}
 
\caption{Experiments on Robustness of loss for continuous treatments. The indicators in the second column denotes whether we use the adversarial training in covariate space (V-GAN) and the reconstruction term for latent features (Z-Rec). Each method was run for five times independently and the standard deviations are shown.}
\label{table:ablation}
\end{table}


\addtolength{\textheight}{-.3in}%

\section{Conclusion}
\label{sec:conc}
In this paper, we developed a novel CausalEGM model, which utilizes the advances in deep generative neural networks for dealing with confounders and estimating the treatment effect in causal inference. CausalEGM enables an efficient encoding, which maps high-dimensional covariates to a low-dimensional latent space. We use GAN-based adversarial training and autoencoder-based reconstruction to guarantee that the latent features are independent of each other and contain the necessary variations in covariates for a good reconstruction. CausalEGM is flexible to estimate the treatment effect for both individuals and populations under either binary or continuous treatment setting. In a series of systematic experiments, CausalEGM demonstrates superior performance over other existing methods.

A number of extensions and refinements of the CausalEGM model are left open. Here, we provide several directions for further exploration. First, although we use GAN-based adversarial training to guarantee the independence in latent features, it is worth trying to incorporate the approximation error in the generation process to analyze the behavior of CausalEGM's convergence. Second, it should be promising to study the complexity of the hyperparameter in CausalEGM when applying to datasets with various sample sizes.


\bigskip





\bibliographystyle{chicago.bst}

\bibliography{Bibliography-MM-MC}
\newpage

\setcounter{page}{1}
\begin{appendices}
\section{Proofs of Theorems and Lemmas}
\paragraph{Proof of Lemma~\ref{thm:identifiable}}
\begin{align*}
\mu(x) &= \mathbb{E}(Y(x)) \\
&= \int\mathbb{E}(f(x,V,\epsilon)|Z_0=z_0)p_{Z_0}(z_0)dz_0 \\
 &= \int\mathbb{E}(f(X,V,\epsilon)|X=x,Z_0=z_0)p_{Z_0}(Z_0)dz_0  \\ 
 &= \int\mathbb{E}(Y|X=x,Z_0=z_0)p_{Z_0}(z_0)dz_0 \\
\end{align*}
where Assumption~\ref{as:modified unconfounderness} is used to obtain the third equality.



\paragraph{Proof of Lemma~\ref{thm:estimation_error_inequality}} 
Using the triangle inequality, we have
\begin{equation}
\left\{
\begin{aligned}
\mathbb{E}_{0}||Y-\hat{f}_M(X,Z_0,Z_1)||_2^2&\le \mathbb{E}_n||Y-\hat{f}_M(X,Z_0,Z_1)||_2^2+|(\mathbb{E}_n-\mathbb{E}_0)(||Y-\hat{f}_M(X,Z_0,Z_1)||_2^2)|\\
&\le\mathbb{E}_n||Y-\hat{f}_M(X,Z_0,Z_1)||_2^2+\sup_{f\in \mathscr{F}_M}|(\mathbb{E}_n-\mathbb{E}_0)(||Y-f(X,Z_0,Z_1)||_2^2)| \\
\mathbb{E}_{0}||X-\hat{h}_M(Z_0,Z_2)||_2^2&\le \mathbb{E}_n||X-\hat{h}_M(Z_0,Z_2)||_2^2+|(\mathbb{E}_n-\mathbb{E}_0)(||X-\hat{h}_M(Z_0,Z_2)||_2^2)|\\
&\le\mathbb{E}_n||X-\hat{h}_M(Z_0,Z_2)||_2^2+\sup_{h\in \mathscr{F}_M}|(\mathbb{E}_n-\mathbb{E}_0)(||X-h(Z_0,Z_2)||_2^2)|\\
\mathbb{E}_{0}||V-\hat{g}_M(\hat{e}_M(V))||_2^2&\le \mathbb{E}_n||V-\hat{g}_M(\hat{e}_M(V))||_2^2+|(\mathbb{E}_n-\mathbb{E}_0)(||V-\hat{g}_M(\hat{e}_M(V))||_2^2)|\\&\le\mathbb{E}_n||V-\hat{g}_M(\hat{e}_M(V))||_2^2+\sup_{g,e\in \mathscr{F}_M}|(\mathbb{E}_n-\mathbb{E}_0)(||V-g(e(V))||_2^2)|\\
d(P_{Z^0},P_{\hat{e}_M(V)};\mathscr{A}_M)&\le d(P_{\hat{e}_M(V)},P_{Z_{emp}};\mathscr{A}_M)+ d(P_{Z_{emp}},P_{Z^0};\mathscr{A}_M)
\end{aligned}
\right.
\end{equation}
Then by the definition of empirical risk minimizer, we further have 

\begin{equation}
\left\{
\begin{aligned}
\mathbb{E}_n||Y-\hat{f}_M(X,Z_0,Z_1)||_2^2&\le \mathbb{E}_n||Y-f^0_M(X,Z_0,Z_1)||_2^2 \\
\mathbb{E}_n||X-\hat{h}_M(Z_0,Z_2)||_2^2&\le \mathbb{E}_n||X-h^0_M(Z_0,Z_2)||_2^2\\
\mathbb{E}_n||V-\hat{g}_M(\hat{e}_M(V))||_2^2&\le \mathbb{E}_n||V-g^0_M(e^0_M(V))||_2^2\\
d(P_{\hat{e}_M(V)},P_{Z_{emp}};\mathscr{A}_M)&\le d(P_{e^0_M(V)},P_{Z_{emp}};\mathscr{A}_M)
\end{aligned}
\right.
\end{equation}

Then using the triangle inequality again, we have
\begin{equation}
\left\{
\begin{aligned}
\mathbb{E}_n||Y-f^0_M(X,Z_0,Z_1)||_2^2&\le\mathbb{E}_0||Y-f^0_M(X,Z_0,Z_1)||_2^2+|(\mathbb{E}_n-\mathbb{E}_0)(||Y-f^0_M(X,Z_0,Z_1)||_2^2)| \\
&\le\mathbb{E}_0||Y-f^0_M(X,Z_0,Z_1)||_2^2+\sup_{f\in \mathscr{F}_M}|(\mathbb{E}_n-\mathbb{E}_0)(||Y-f(X,Z_0,Z_1)||_2^2)\\
\mathbb{E}_n||X-h^0_M(Z_0,Z_2)||_2^2&\le \mathbb{E}_0||X-h^0_M(Z_0,Z_2)||_2^2 + |(\mathbb{E}_n-\mathbb{E}_0)(||X-h^0_M(Z_0,Z_2)||_2^2)|\\
&\le\mathbb{E}_0||X-h^0_M(Z_0,Z_2)||_2^2+\sup_{h\in \mathscr{F}_M}|(\mathbb{E}_n-\mathbb{E}_0)(||X-h(Z_0,Z_2)||_2^2)|\\
\mathbb{E}_n||V-g^0_M(e^0_M(V))||_2^2&\le \mathbb{E}_0||V-g^0_M(e^0_M(V))||_2^2 + |(\mathbb{E}_n-\mathbb{E}_0)(||V-g^0_M(e^0_M(V))||_2^2)|\\
&\le\mathbb{E}_0||V-g^0_M(e^0_M(V))||_2^2+\sup_{g,e\in \mathscr{F}_M}|(\mathbb{E}_n-\mathbb{E}_0)(||V-g(e(V))||_2^2)|\\
d(P_{e^0_M(V)},P_{Z_{emp}};\mathscr{A}_M)&\le d(P_{e^0_M(V)},P_{Z^0};\mathscr{A}_M)+d(P_{Z_{emp}},P_{Z^0};\mathscr{A}_M)
\end{aligned}
\right.
\end{equation}

Combining all terms above, we can then get the desired results.

\paragraph{Proof of theorem~\ref{thm: Estimation_Error}}
Each of $\alpha_{M,n}, \beta_{M,n}$ and $\zeta_{M,n}$ can be upper bounded by $$2\sup_{F\in \mathscr{F}_M}|(\mathbb{E}_n-\mathbb{E}_0)F|=2\sup_{F\in \mathscr{F}_M}|{\frac{1}{n}\sum_{i=1}^{n}F(O_i)-\mathbb{E}_0[F(O)]}|.$$
For $\gamma_{M,n}$, note if $\mathscr{D}_M$ is the class of the binary discriminator networks that classifies the class of the measurable sets $\mathscr{A}_M$, let $Z_1,...,Z_n$ be the i.i.d. samples from $Z_0$, $Z \sim Z_0,$ we have
$$ d(P_{Z_{emp}},P_{Z^0};\mathscr{A}_M)= \sup_{D\in \mathscr{D}_M} |\frac{1}{n}\sum_{i=1}^{n}\mathds{1}_{\{D(Z_i)=1\}}-\mathbb{E}_0[\mathds{1}_{\{D(Z)=1\}}]|.$$
Hence $\gamma_{M,n}$ can be upper bounded by $\sup_{D\in \mathscr{D}_M}|(\mathbb{E}_n-\mathbb{E}_0)D|$. Now for any \textit{$b$-uniformly bounded} function class $\mathscr{F}$, the uniform law of large numbers states that for all $n\geq1$ and $\delta\geq0$, we have:
$$\sup_{f\in \mathscr{F}}|(\mathbb{E}_n-\mathbb{E}_0)f|\leq2\mathscr{R}_n(\mathscr{F})+\delta.$$
with probability at least $1-2e^{-\frac{n\delta^2}{8b^2}}$. We then complete the proof by Assumption~\ref{as:uniformly_bounded}  and by applying this bound to the combined terms of $\alpha_{m,n}, \beta_{m,n}, \gamma_{m,n}$ and $\zeta_{m,n}:$
\begin{equation}
\begin{aligned}
&\mathbb{P}(R^0(\hat{f}_M,\hat{h}_M, \hat{e}_M,\hat{g}_M)-\inf_{f,h,e,g\in \mathscr{F}_M}R^0(f,h,e,g)\leq12\mathscr{R}_n(\mathscr{F}_M)+ 4\mathscr{R}_n(\mathscr{D}_{M})+\delta)\geq \\
&\mathbb{P}(\alpha_{M,n}+\beta_{M,n}+\gamma_{M,n}+\zeta_{M,n}\leq 12\mathscr{R}_n(\mathscr{F}_M)+ 4\mathscr{R}_n(\mathscr{D}_{M})+\delta)\geq \\
&\mathbb{P}(6\sup_{F\in \mathscr{F}_M}|(\mathbb{E}_n-\mathbb{E}_0)F|+2\sup_{D\in \mathscr{D}_M}|(\mathbb{E}_n-\mathbb{E}_0)D|\leq12\mathscr{R}_n(\mathscr{F}_M)+ 4\mathscr{R}_n(\mathscr{D}_{M})+\delta))\geq \\
&\mathbb{P}(\sup_{F\in \mathscr{F}_M}|(\mathbb{E}_n-\mathbb{E}_0)F|\leq 2\mathscr{R}_n(\mathscr{F}_M)+\frac{\delta}{8}, \sup_{D\in \mathscr{D}_M}|(\mathbb{E}_n-\mathbb{E}_0)D|\leq 2\mathscr{R}_n(\mathscr{D}_{M})+\frac{\delta}{8})\geq \\
&1-\mathbb{P}(\sup_{F\in \mathscr{F}_M}|(\mathbb{E}_n-\mathbb{E}_0)F|> 2\mathscr{R}_n(\mathscr{F}_M)+\frac{\delta}{8})-\mathbb{P}(\sup_{D\in \mathscr{D}_M}|(\mathbb{E}_n-\mathbb{E}_0)D|>2\mathscr{R}_n(\mathscr{D}_{M})+\frac{\delta}{8})\geq \\
&1-4e^{-\frac{n\delta^2}{512b^2}}
\end{aligned}
\end{equation} 
where we use the result of the uniform law of large numbers in the last inequality.



\paragraph{Proof of theorem~\ref{thm: consistency}}

By Theorem~\ref{thm: Estimation_Error} for all $\delta'>0$, we would have
$$
(\dagger): R^0({f^{\ast},h^{\ast},e^{\ast},g^{\ast}})\leq R^0(f,h,e,g)+\delta'
$$
a.s. for any $(f,h,e,g)$ that can be approximated by neural networks. In particular, we can choose $\delta'$ to be the same $\delta$ as in Assumption~\ref{as:reconstruction_error}. Using Equations (\ref{eqn: basic_eqn}) we have for any $(f,h,e,g)$,
\begin{align*}
R^0(f,h,e,g)&=\mathbb{E}_{0}||(f^0-f)(X,Z_0,Z_1)||_2^2 +\sigma^2_1+\mathbb{E}_{0}||(h^0-h)(Z_0,Z_2)||_2^2+\sigma^2_2 \\
&+d(P_{Z^0},P_{e(V)};\mathscr{A}_M)+\mathbb{E}_{0}||(V-g(e(V)))||_2^2
\end{align*}

Now suppose 
$$e_3', g' = \underset{e_3, g}{\mathrm{argmin}}\{d(P_{Z^0},P_{(e_0^0, e_1^0, e_2^0, e_3)(V))};\mathscr{A}_m) + \mathbb{E}_{0}||V-g(e_0^0, e_1^0, e_2^0, e_3)(V))||_2^2\}$$
Substituting
$$
f=f^0, h=h^0, e=(e_0, e_1^0, e_2^0, e_3') \text{ and } g=g'
$$
into the right-hand side of $(\dagger)$ gives:

\begin{align*}
R^0({f^{\ast},h^{\ast},e^{\ast},g^{\ast}}) &\leq \sigma^2_1+\sigma^2_2+d(P_{Z^0},P_{(e_0^0, e_1^0, e_2^0, e_3')(V))};\mathscr{A}_M)+\mathbb{E}_{0}||V-g'((e_0^0, e_1^0, e_2^0, e_3')(V))||_2^2+\delta \\
& \leq  \sigma^2_1+\sigma^2_2+d(P_{Z^0},P_{(e_0^0, e_1^0, e_2^0, \tilde{e_3})(V))};\mathscr{A}_M)+\mathbb{E}_{0}||V-\tilde{g}((e_0^0, e_1^0, e_2^0, \tilde{e}_3)(V))||_2^2+\delta \\
&\overset{Asm.~\ref{as:reconstruction_error}}{\leq}\sigma^2_1 + \sigma^2_2 + \mathbb{E}_{0}||V-g^{\ast}(e^{\ast}(V))||_2^2 + 2\delta
\end{align*}
On the other hand, we have 

\begin{equation}
\begin{aligned}
R^0({f^{\ast},h^{\ast},e^{\ast},g^{\ast}})=&\mathbb{E}_{0}||(f^0-f^{\ast})(X,Z_0,Z_1)||_2^2+\sigma^2_1+\mathbb{E}_{0}||(h^0-h^{\ast})(Z_0,Z_2)||_2^2+\sigma^2_2+\\
&d(P_{Z^0},P_{e^{\ast}(V)};\mathscr{A}_M)+\mathbb{E}_{0}||(V-g^{\ast})(e^{\ast}(V))||_2^2
\end{aligned}
\end{equation}

combining with above, we then get the desired inequality.

\section{Simulation example for assumption~\ref{as:reconstruction_error}}\label{sim_asm4}

Assumption~\ref{as:reconstruction_error} is expected to hold with a small $\delta$ and the dimension of the covariates $V$ can be effectively reduced. This is illustrated by the following simulation study. Let $V\in \mathcal{V}$ be a continuous random variable that follows a multivariate Gaussian distribution $V\sim N(\bm{\mu}, \bm{\Sigma})$ where $\bm{\mu}\in \mathbb{R}^p$ and $\bm{\Sigma}\in \mathbb{R}^{p\times p}$. We aim to find encoding function $e$ and generative/decoder function $g$, which follow the mappings $e:\mathcal{V}\rightarrow \mathbb{R}^q$ and $g:\mathbb{R}^q\rightarrow \mathcal{V}$ where $q \ll p$. First, we factorize the covariance matrix as   
\begin{equation}
\label{covariance_matrx}
\begin{aligned}
\bm{\Sigma}=\bm{U}\bm{\Lambda}\bm{U}^T
\end{aligned}
\end{equation} 
where the columns of $\bm{U}$ form the eigenvectors associated with the eigenvalues in diagonal elements of $\bm{\Lambda}$. We further sort all the eigenvalues in descending order as $\bm{\Lambda}=diag(\lambda_1,...,\lambda_p)$ where $\lambda_i\ge\lambda_j$ for any $i<j$.

By linear transformation, it is easily proven that $\bm{T}=(\bm{U}\bm{\Lambda}^{\frac{1}{2}})^{-1}(\bm{V}-\bm{\mu})$ follows a standard multivariate Gaussian distribution where $\bm{T}\sim N(\bm{0},\bm{I})$. This linear transformation could be considered as the underlying encoding function where a standard Gaussian distribution is present in the latent space. In the dimension reduction scenario, it is expected that a small fraction of eigenvalues in $\bm{\Sigma}$ could explain the majority of the total variation in $V$. So we design the following generating process. 

We set $p=50$, $q=13$, and the diagonal elements of $\bm{\Lambda}$ to be
\begin{equation}
\begin{aligned}
\lambda_i=
\begin{cases}
5-\frac{1}{9}(i-1), & i\le10,\\
0.1-\frac{1}{400}(i-11), & 11\le i\le50
\end{cases}
\end{aligned}
\end{equation}
where the first $13$ principle components can explain 95.96\% of the variation contained in $V$. To generate $V$, the mean vector $\bm{\mu}$ is sampled from a uniform distribution ${\mu}_i\sim U(-1,1)$, the covariance matrix $\bm{\Sigma}$ is constructed by Equation (\ref{covariance_matrx}) where the columns of $\bm{U}$ are a set of random orthonormal basis. To construct the features from $V$ for predicting treatment $X$ and outcome $Y$, we set the three components $e^0_0(V)$, $e^0_1(V)$, and $e^0_2(V)$ in the encoder network $e$ as follows
\begin{equation}
\left\{
\begin{aligned}
e^0_0(V)=&(t_8(V)+t_{11}(V))/\sqrt2 \\
e^0_1(V)=&(t_9(V)+\sum_{i=12}^{20}t_i(V))/\sqrt{10} \\
e^0_2(V)=&(t_{10}(V)+\sum_{i=22}^{30}t_i(V))/\sqrt{10} \\
\end{aligned}
\right.
\end{equation}
where $t_i(V)$ denotes that $i^{th}$ element of the linear transformation $\bm{T}=(\bm{U}\bm{\Lambda}^{\frac{1}{2}})^{-1}(\bm{V}-\bm{\mu})$. It is easily proven that $e^0_k(V)\sim N(0,1)$ for $k\in \{1,2,3\}$, which satisfies the condition \ref{eqn:multinormal}. The treatment and outcome can then be generated based on the features of $V$ as 
\begin{equation}
\left\{
\begin{aligned}
Y=&f(X,e^0_0(V), e^0_1(V))+\epsilon_1\\
X=&h(e^0_0(V), e^0_2(V))+\epsilon_2
\end{aligned}
\right.
\end{equation}
where $e^0_0(V)$, $e^0_1(V)$, and $e^0_2(V)$ can be considered as the constructed features from $V$ for predicting $X$ and $Y$. For implementation, we set the first three parts of encoder $e$ network to be the fixed functions, $e^0_0(\cdot)$, $e^0_1(\cdot)$, and $e^0_2(\cdot)$. The fourth part of encoder $e$ is trainable, which is set to be 10-dimensional. According to the Principal Component Analysis (PCA) \cite{pearson1901liii}, the theoretical optimal reconstruction error using a $q$-dimensional feature is
\begin{equation}
\label{threotical_error}
\begin{aligned}
\mathcal{L}_{rec}=\sum_{i=q+1}^{p}\lambda_i
\end{aligned}
\end{equation} 

\begin{figure}[htbp]
  \centering
  \includegraphics[width=1.0\columnwidth]{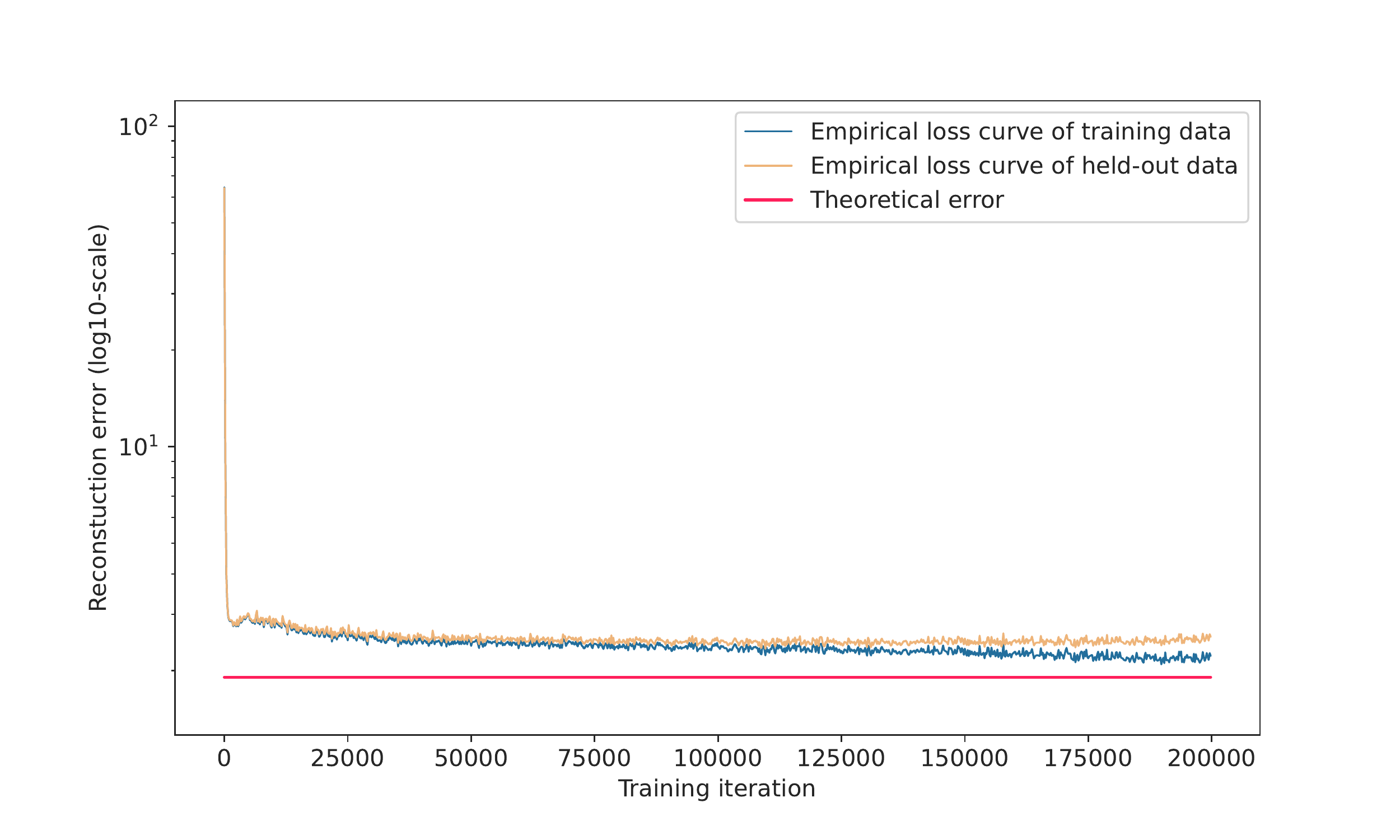}
  \caption{The simulation experiment for verifying the assumption \ref{as:reconstruction_error}}
  \label{Assump4_sim}
\end{figure}

In the above simulation example, $\mathcal{L}_{rec}=1.907$. Then we generate $N=50000$ $i.i.d$ samples of $V$, and then use the data to train the above partially fixed encoder-decoder. To avoid overfitting of neural nets, we additionally generate $10000$ hold-out samples of $V$. As shown in Figure \ref{Assump4_sim}, the empirical reconstruction error of the held-out data reaches the minimum ($2.339$) at iteration $109600$. In this simulation, $\delta$ in assumption \ref{as:reconstruction_error} can be as small as $0.432$, which only occupies less than  $1\%$ of all variation contained in $V$ ($\sum_{i=1}^{p}\lambda_i$).

\section{ACIC 2018 data details}
\label{app_data}
ACIC 2018 data were constructed from Linked Births and Infant Deaths Database (LBIDD), which provide a valuable resource for evaluating the performance in estimating treatment effects. We chose the most comprehensive datasets by selecting the highest degree for the generation function (e.g, outcome generation function). The highest degrees of outcome generation function in datasets with sample size 1k, 10k, 50k are 98, 101, and 75 respectively. The nine datasets used in this study are summarized in Table \ref{table:data}.

\begin{table}
  \centering

  \resizebox{\textwidth}{!}{
  \begin{tabular}{cccccl}
    \toprule
    Sample size&Ufid&Treated percentage&True ATE\\
    \midrule
    \multirow{3}*{Datasets-1k}  
    &629e3d2c63914e45b227cc913c09cebe&36.76\%&0.006208\\
    &35524a031525484dab3b06f3728c708e&10.01\%&0.513605\\
    &a957b431a74a43a0bb7cc52e1c84c8ad&46.44\%&6.373162\\
                        
    \midrule
    \multirow{3}*{Datasets-10k}                 
    &71f29913f174456e9fe2727b1b86b8b3&57.98\%&9.243389\\
    &fda655aeb8644c9db5c543ed9d1006ad&22.05\%&-0.0566\\
    &05fdeea9fcb64b3885e6ebfb85b4ce90&12.08\%&0.025841\\
    \midrule
    \multirow{3}*{Datasets-50k} 
    &1c565ac309074f178a377c2759333209&14.48\%&-0.468357	\\
    &b73beac2f4c349fb981880399d4c88a6&18.79\%&-0.049168\\
    &d5bd8e4814904c58a79d7cdcd7c2a1bb&54.50\%&-0.296505\\
   
  \bottomrule
\end{tabular}}
 
\caption{Details of binary treatment datasets used in this study. Each dataset has a unique ufid series number.}
\label{table:data}
\end{table}

\section{Robustness and scalability analysis}
\label{app_scala}
\begin{figure}[htbp]
  \centering
  \includegraphics[width=1.\columnwidth]{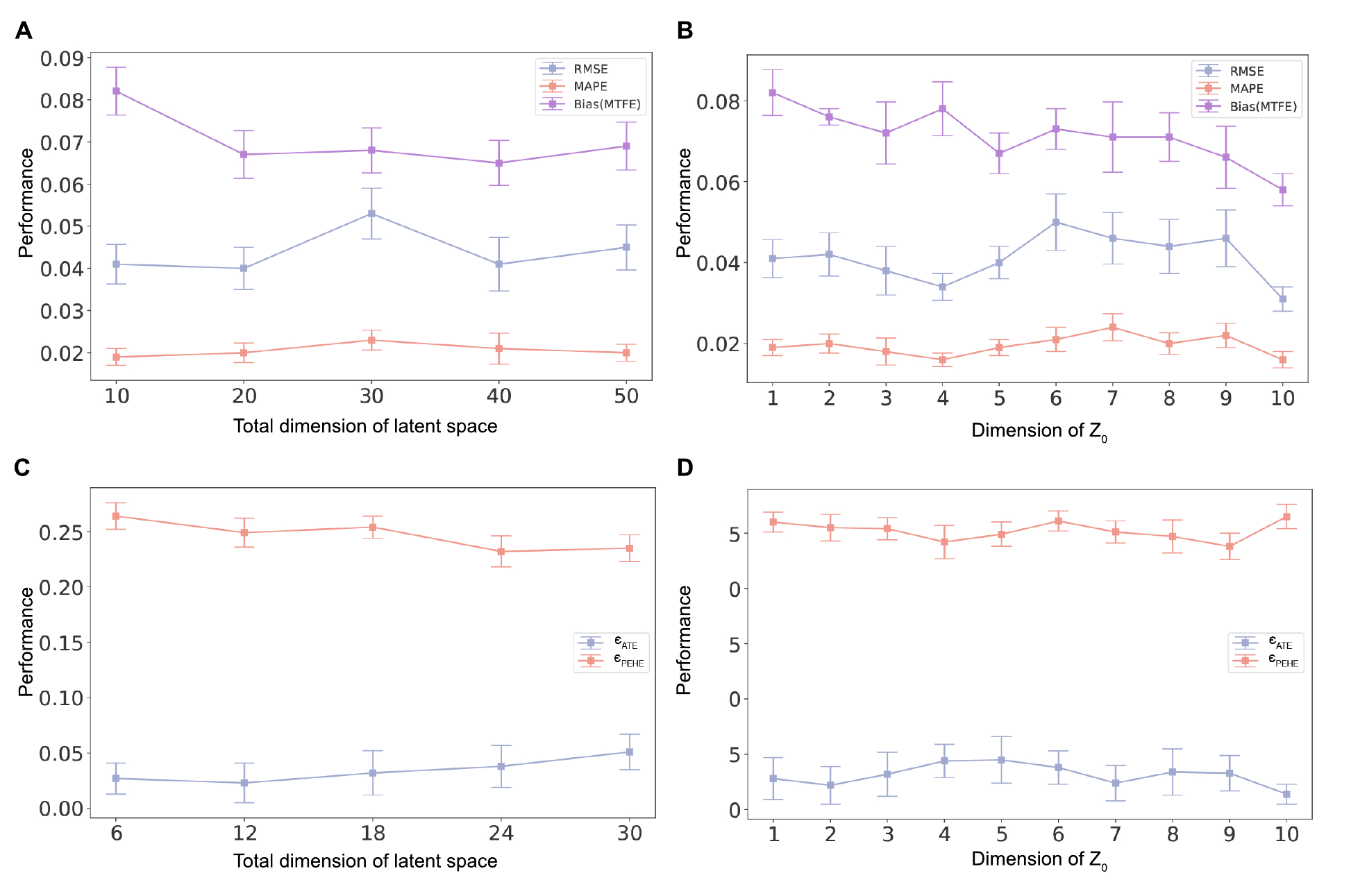}
  \caption{Robustness of latent dimensions in CausalEGM model. The error bar denotes the standard deviation of 10 independent runs. (A) Performance when the total dimension of latent features varies from 10 to 50 in Hiranos and Imbens dataset. (B) Performance when the dimension of $Z_0$ varies from 1 to 10 while dimensions of other $Z_i (i=1,2,3)$ are fixed in Hiranos and Imbens dataset. (C) Performance when the total dimension of latent features varies from 6 to 30 in a LBIDD dataset. (D) Performance when the dimension of $Z_0$ varies from 1 to 10 while dimensions of other $Z_i (i=1,2,3)$ are fixed in a LBIDD dataset.}
  \label{fig:robustness}
\end{figure}

To demonstrate the robustness and scalability of CausalEGM, we designed a series of experiments as follows. For the robustness analysis, it is important to evaluate how the dimension for latent features $Z$ will affect the performance of CausalEGM model. We focus on both the total dimension of latent feature $Z$ and also the dimension of the common latent features $Z_0$ that affect both treatment and outcome. For continuous experiments, we choose Hiranos and Imbens dataset for example. On the one hand, the dimension for $Z_0,Z_1,Z_2,Z_3$ is set to be $\{(k,k,k,7k)\|k=1,2,...,5\}$ where $k=1$ is used as default in the main result. On the other hand, we set the dimension for $Z_0$ to be ranging from 1 to 10 while dimensions of other latent features ($Z_i (i=1,2,3)$) are fixed to $(1,1,7)$, respectively. It is noted that the performance has a small fluctuation by varying the dimension of either total latent features or only common latent features (Figure \ref{fig:robustness}A-B). We use similar
settings in the binary experiment for robustness analysis. We chose a dataset from LBIDD with a sample size equal to 1000 for instance. On the one hand, the dimension for $Z_0,Z_1,Z_2,Z_3$ is set to be $\{(k,k,2k,2k)|k=1,2,...,5\}$ where $k=3$ is used as default. On the other hand, we set the dimension for $Z_0$ to be ranging from 1 to 10 while dimensions of other latent features ($Z_i (i=1,2,3)$) are fixed to $(3,6,6)$. It is observed that the performance does not change significantly by varying the dimension of latent features (Figure \ref{fig:robustness}C-D). Such experiments in both continuous and binary treatment settings demonstrate the robustness of CausalEGM in terms of choosing the latent dimensions.

\begin{figure}[htbp]
  \centering
  \includegraphics[width=1.\columnwidth]{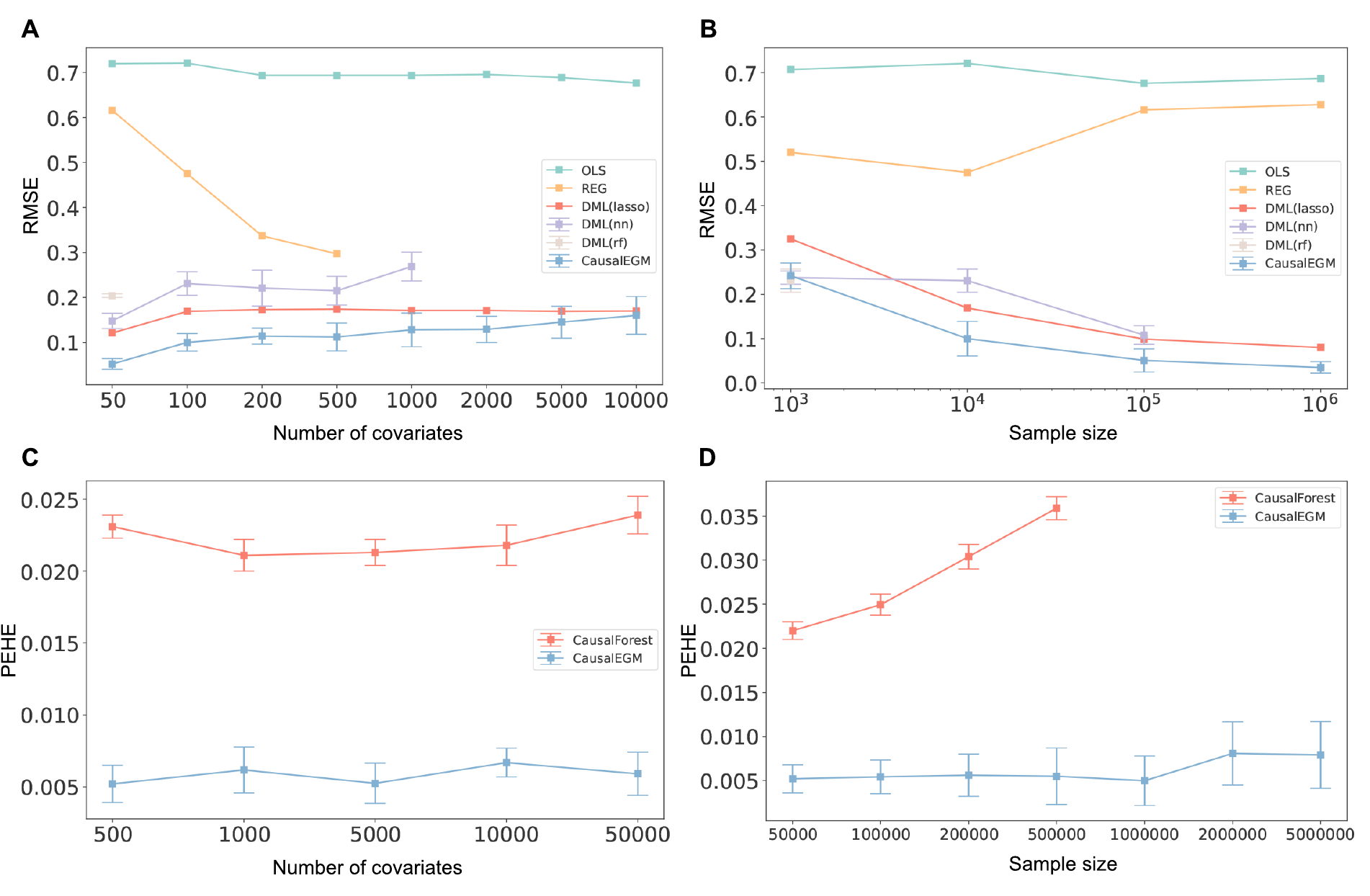}
  \caption{Scalability of CausalEGM model in terms of covariates and sample size. The error bar denotes the standard deviation of 10 independent runs. (A) Performance when changing the number of covariates from 50 to 10000 in Hirano and Imbens dataset. (B) Performance when changing the sample size from $10^3$ to $10^6$ in Hirano and Imbens dataset. (C) Performance when changing the number of covariates from 500 to 50000 in a LBIDD dataset. (D) Performance when changing the sample size from $50000$ to $5000000$ in the LBIDD dataset.}
  \label{fig:scalability}
\end{figure}

For the scalability analysis, we are interested in 1) whether CausalEGM can handle datasets with large sample sizes; and 2) whether CausalEGM can handle datasets with a large number of covariates. We designed the following experiments to test the scalability of CausalEGM. For the continuous treatment experiment, we selected Hirano and Imbens dataset. We first change the number of covariates from 50 to 10000 while the sample size is 10000. Note that only OLS, DML(lasso) and CausalEGM can handle covariates more than 1000 while other comparison methods failed (Figure \ref{fig:scalability} A). Next, we fix the number of covariates to 100 while changing the sample size from $10^3$ to $10^6$. Only OLS, REG, DML(lasso), and CausalEGM are able to handle large sample size $10^6$ (Figure \ref{fig:scalability} B). Except for a small sample size situation (e.g., 1000) where CausalEGM achieves comparable performance compared to DML(nn) and DML(rf), CausalEGM consistently outperforms all comparison methods by either changing the number of covariates or sample size. Similarly, for the binary treatment experiment, we chose one of the largest dataset from LBIDD with a sample size equal to 50000. For such a semi-synthetic dataset, we increase the number of covariates by adding new covariates that are linear combinations of existing covariates where the combination coefficients follow the standard normal distribution. We increase the sample size by augmenting the data by randomly repeating the existing samples. We tested the performance of CausalEGM and the best baseline method CausalForest. We first change the number of covariates from 500 to 50000 while the sample size is 50000. Note that the performance of CausalForest first increases a little and then decreases while CausalEGM is generally more robust when changing the number of covariates (Figure \ref{fig:scalability} C). Next, we fix the number of covariates to the original 177 while changing the sample size from $50000$ to $5000000$. Note that CausalForest failed when the sample size increases to 1 million while CausalForest is capable of handling extremely large datasets with more than 5 million samples (Figure \ref{fig:scalability}D). Note that we benchmarked all methods using the Stanford Sherlock computing cluster where the memory usage for each method is limited to 50 GB and running time is limited to 7 days in the scalability experiments. To sum up, CausalEGM can handle significantly larger datasets than CausalForest.

\end{appendices}

\newpage

\end{document}